\documentclass[10pt,twocolumn,letterpaper]{article}

\usepackage{iccv}
\usepackage{times}
\usepackage{epsfig}
\usepackage{graphicx}
\usepackage{amsmath}
\usepackage{amssymb}
\usepackage{caption}
\usepackage{multirow}

\usepackage[breaklinks=true,bookmarks=false]{hyperref}

\iccvfinalcopy 

\def\iccvPaperID{10888} 

\ificcvfinal\fi

\begin{document}

\title{Towards High-Fidelity Text-Guided 3D Face Generation and Manipulation Using only Images}


\author{Cuican Yu\textsuperscript{*1}, Guansong Lu\textsuperscript{*2},
Yihan Zeng\textsuperscript{*2},
Jian Sun\textsuperscript{1},
Xiaodan Liang\textsuperscript{3},\\
Huibin Li\textsuperscript{1},
Zongben Xu\textsuperscript{1},
Songcen Xu\textsuperscript{2},
Wei Zhang\textsuperscript{2},
Hang Xu\textsuperscript{2$\dagger$}\\
{\textsuperscript{1} Xi'an Jiaotong University}
{\textsuperscript{2} Huawei Noah's Ark Lab}
{\textsuperscript{3} Sun Yat-sen University}\\
{\tt\small ccy2017@stu.xjtu.edu.cn}
{\tt\small \{luguansong,zengyihan2,xusongcen,wz.zhang\}@huawei.com}\\
{\tt\small \{jiansun,huibinli,zbxu\}@xjtu.edu.cn}
{\tt\small \{xdliang328,chromexbjxh\}@gmail.com}}


\twocolumn[{
\renewcommand\twocolumn[1][]{#1}
\maketitle
\begin{center}
    \captionsetup{type=figure}
    \includegraphics[width=1\textwidth]{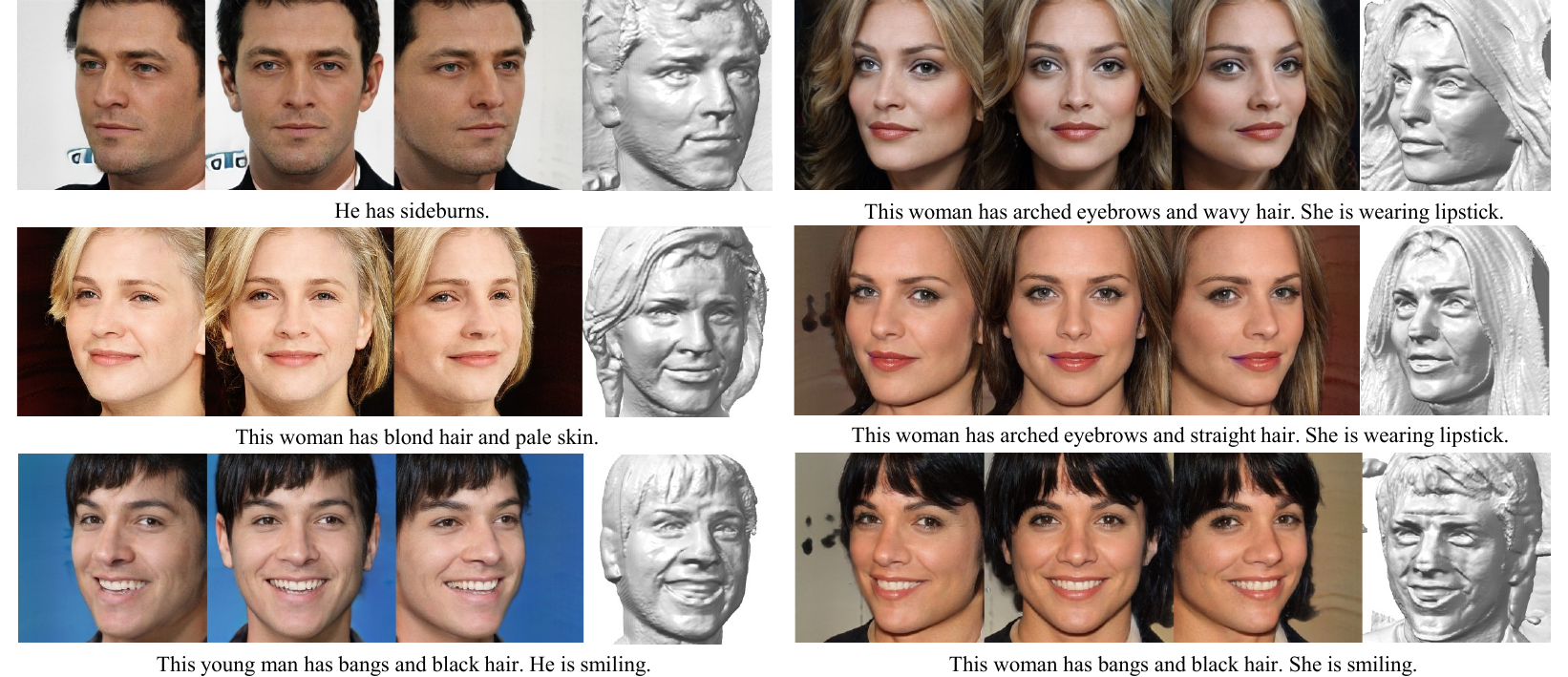}
    \captionof{figure}{3D face generations of our TG-3DFace. Given input texts, TG-3DFace can generate high-quality 3D faces with multi-view-consistent rendered face images and detailed 3D face meshes. Notably, fine-grained facial attributes are well controlled by the input texts.}
    \label{fig:main-result}
\end{center}
}]


\renewcommand{\thefootnote}{}
\footnotetext{$*$ Equal contribution}
\footnotetext{$\dagger$ Corresponding author }

\ificcvfinal\thispagestyle{empty}\fi

\begin{abstract}
   Generating 3D faces from textual descriptions has a multitude of applications, such as gaming, movie, and robotics. Recent progresses have demonstrated the success of unconditional 3D face generation and text-to-3D shape generation. However, due to the limited text-3D face data pairs, text-driven 3D face generation remains an open problem. In this paper, we propose a text-guided 3D faces generation method, refer as TG-3DFace, for generating realistic 3D faces using text guidance.
Specifically, we adopt an unconditional 3D face generation framework and equip it with text conditions, which learns the text-guided 3D face generation with only text-2D face data. On top of that, we propose two text-to-face cross-modal alignment techniques, including the global contrastive learning and the fine-grained alignment module, to facilitate high semantic consistency between generated 3D faces and input texts. Besides, we present directional classifier guidance during the inference process, which encourages creativity for out-of-domain generations.
Compared to the existing methods, TG-3DFace creates more realistic and aesthetically pleasing 3D faces, boosting 9\%  multi-view consistency (MVIC) over Latent3D.
The rendered face images generated by TG-3DFace
achieve higher FID and CLIP score than text-to-2D face/image generation models,
demonstrating our superiority in generating realistic and semantic-consistent textures.
\end{abstract}

\section{Introduction}
 3D Face generation is a critical technology with diverse applications in various industry scenarios, e.g., movies and games. Recent works have demonstrated the success of 3D face generation with image reconstruction ~\cite{lattas2020avatarme, lattas2021avatarme++} and unconditional generation methods \cite{pigan,athar2022rignerf,sun2022fenerf,moschoglou20203dfacegan,eg3d}. Despite the photo-realistic 3D face results, the generation process cannot be guided by texts, which has the potential to increase creativity and efficiency. Therefore it is highly demanded to take a step toward text-guided 3D face generation.

Existing methods have been explored to generate 3D shapes and human bodies based on given texts~\cite{chen2018text2shape,liu2022implicit-text-3D-generation,sanghi2021clip-forge,hong2022avatarclip,text2mesh}, which enables the controllable generation under text guidance. However, it is not feasible to directly apply those generation methods for 3D face generation, owing to two fact:1) The lack of large-scale text-3D face data pairs for model training. 2) The richness of 3D facial attributes that contains much more geometrical details than common 3D objects. Though recent works~\cite{canfes2022text,aneja2023clipface}  make attempts to semantically manipulate the shape or texture of 3D faces to boost 3D face generation results, they still lead to results with poor realism and aesthetic appeal such as the loss of hair, which limits the practical applications. Based on the above observation, it requires a rethink of a fine-grained text-driven 3D face generation framework.

  To address the above issues, we present a novel fine-grained text-driven 3D face generation framework, named TG-3DFace, to generate high-quality 3D faces that are semantically consistent with the input texts.
 Specifically, TG-3DFace contains a text-conditioned 3D generation network and two text-to-face cross-modal alignment techniques. Firstly, we adopt the architecture design of EG3D \cite{eg3d}, which is an unconditional 3D shape generative adversarial network, and learn 3D shape generation from single-view 2D images. We inject the texture condition into the generator and discriminator networks to enable 3D face generation under the guidance of input texts. Such text-guided 3D face generative model can thus conduct training on text-2D face images instead of text-3D face shapes, enabling to transfer the semantic consistency between texts and 2D face images to guide 3D face generation.
Besides, considering the richness of fine-grained facial attributes that increases the difficulty of aligning 3D face and input texts, we design two text-to-face cross-modal alignment techniques, including global text-to-face contrastive learning and fine-grained text-to-face alignment module.
The text-to-face contrastive learning aligns the features of the rendered face images with their paired text and maximizes the distance between the unpaired ones in the embedding space, which facilitates global semantic consistency.
The fine-grained text-to-face alignment is designed to align the part-level facial features of the rendered face image to the part-level text features, to achieve fine-grained semantic alignment between the texts and the generated 3D faces.

Additionally, we utilize the directional vector in the CLIP embedding space, calculated between the input text and the training style prompt text, as an optimization direction to fine-tune the generator for several steps during inference.
In this way, our TG-3DFace can synthesize novel-style face that is never seen during training, such as ``a Pixar-style man".

We evaluate our model on the Multi-Modal CelebA-HQ \cite{xia2021tedigan}, CelebAText-HQ \cite{sea-t2f} and FFHQ-Text \cite{FFHQ_text} datasets. The experimental results and ablation analysis demonstrate that our method can generate high-quality and semantic-consistent 3D faces given input texts. Besides, our method can be applied to downstream applications including single-view 3D face reconstruction and text-guided 3D face manipulation.
In brief, our contributions can be summarized as follows:
\begin{itemize}
    \item We propose a novel 3D face generation framework, TG-3DFace, which equips the unconditional 3D face generation framework with text conditions to generate 3D faces with the guidance of input texts.
    \item We propose two text-to-face cross-modal alignment techniques, including global contrastive learning and fine-grained text-to-face alignment mechanism, which boosts the semantic consistency of generations.
    \item Quantitative and qualitative comparisons confirm that 3D faces generated by our TG-3DFace are more realistic and achieve better semantic consistency with the given textual description.
\end{itemize}

\section{Related Work}
\label{sec_related_work}

\subsection{3D Face Generation}
3DFaceGAN \cite{moschoglou20203dfacegan} applies the generative adversarial networks (GANs \cite{goodfellow2014gan}) to represent, generate and translate 3D facial shapes meshes.
pi-GAN \cite{pigan} presents a SIREN network as the generator to represent the implicit radiance field, which conditioned on an input noise. The authors also propose a mapping network with FiLM conditioning and a progressive growing discriminator to achieve high quality results. AvatarMe \cite{lattas2020avatarme,lattas2021avatarme++} explores to reconstruct photo-realistic 3D faces from a single ``in-the-wild'' face image based on the state-of-the-art 3D texture and shape reconstruction method.
RigNeRF \cite{athar2022rignerf} uses a 3DMM-guided deformable neural radiance field to generate a human portrait trained on a short portrait video. FENeRF \cite{sun2022fenerf} proposes to condition a NeRF generator on decoupled shape and texture latent code to learn the semantic and texture representations simultaneously, which helps to generate more accurate 3D geometry.
EG3D \cite{eg3d} proposes a tri-plane-based hybrid explicit-implicit 3D representation with a high computational efficiency.
They also introduce a dual-discriminator training strategy to enforce the view-consistency of the final output. In contrast to these works, our method explores text-conditioned 3D face generation with other mechanisms to enforce the semantic consistency between the given text and generated 3D face so that the generated 3D faces can be flexibly controlled by the inputted texts.

\subsection{Text-to-Image Generation}
Given a text description, text-to-image generation aims to generate an image to visualize the context described by the text.
There are numbers of works for text-to-image generation along with the generative models, including generative adversarial networks (GANs \cite{goodfellow2014gan}) \cite{GAN-INT-CLS,zhang2017stackgan,zhang2018stackgan++,xu2018attngan,li2019controllable,sisgan,zhu2019dmgan,tao2020dfgan,ye2021xmcgan}, auto-regressive model \cite{vaswani2017attention,dalle,ding2021cogview,esser2021imagebart,ding2022cogview2,zhang2021m6-ufc,lee2022rq-vae,parti} and diffusion model \cite{ho2020ddpm,nichol2021glide,ramesh2022dalle2,imagen,latent-diffusion}.
There are also works focusing on facial image generation. Text2FaceGAN \cite{nasir2019text2facegan} explores to apply the state-of-the-art GAN at the time on text-to-face generation. Stap \etal \cite{stap_Conditional} proposes textStyleGAN to generate facial image from text by conditioning the StyleGAN \cite{stylegan} model on text and then manipulate the generated image in the disentangled latent space to make the result semantically more close to the text.
SEA-T2F \cite{sea-t2f}
presents a Semantic Embedding and Attention network for multi-caption text-to-face generation.
TTF-HD \cite{faces-a-la-carte}, TediGAN \cite{xia2021tedigan} and AnyFace \cite{anyface} propose to align/manipulate the input
latent vector of a pretrained StyleGAN model guided by input text to achieve text-to-face generation/manipulation.
Recently, PixelFace \cite{pixelface} and OpenFaceAN \cite{openfacegan} are proposed to achieve higher performance than StyleGAN-based methods.
However, these works only generate single view images and do not consider 3D face generation, while our method outputs high-quality 3D face shapes and multi-view consistent images.

\subsection{Text-to-3D Shape Generation}

Text2shape \cite{chen2018text2shape} describes a text-conditioned Wasserstein GAN for generating voxelized 3D objects.
CLIP-forge \cite{sanghi2021clip-forge} trains a normalizing flow network to generate 3D shape embedding conditioned on CLIP image embedding
and uses it to generate 3D shapes conditioned on CLIP text embeddings.
Liu \etal \cite{liu2022implicit-text-3D-generation} uses an implicit occupancy representation and proposes a cyclic loss to enforce the consistency
between the generated 3D shape and the input text.
ShapeCrafter \cite{fu2022shapecrafter} explores recursive text-conditioned 3D shape generation that continuously evolve as phrases are added.
However, these works cannot generate realistic 3D objects with high fidelity.
Text2mesh \cite{text2mesh}, CLIP-mesh \cite{khalid2022clip-mesh} and Dreamfields \cite{jain2021dreamfields} use different 3D representations and optimize them with CLIP semantic loss coupled with some regularization terms. DreamFusion \cite{poole2022dreamfusion} instead uses a pretrained 2D text-to-image diffusion model and introduces a novel loss based on probability density distillation to optimize the 3D model.
Some works \cite{hong2022avatarclip,hu2021text,canfes2022text} explore 3D avatar generation and animation from text. However, these works cannot generate high-quality 3D faces as 3D faces contain more details than other 3D shapes (like chair, sofa) and human bodies. On the contrary, our method can generate high-quality 3D faces with rich facial attributes described by the given text.


\begin{figure*}[!htbp]
\begin{center}
\includegraphics[width=0.95\linewidth]{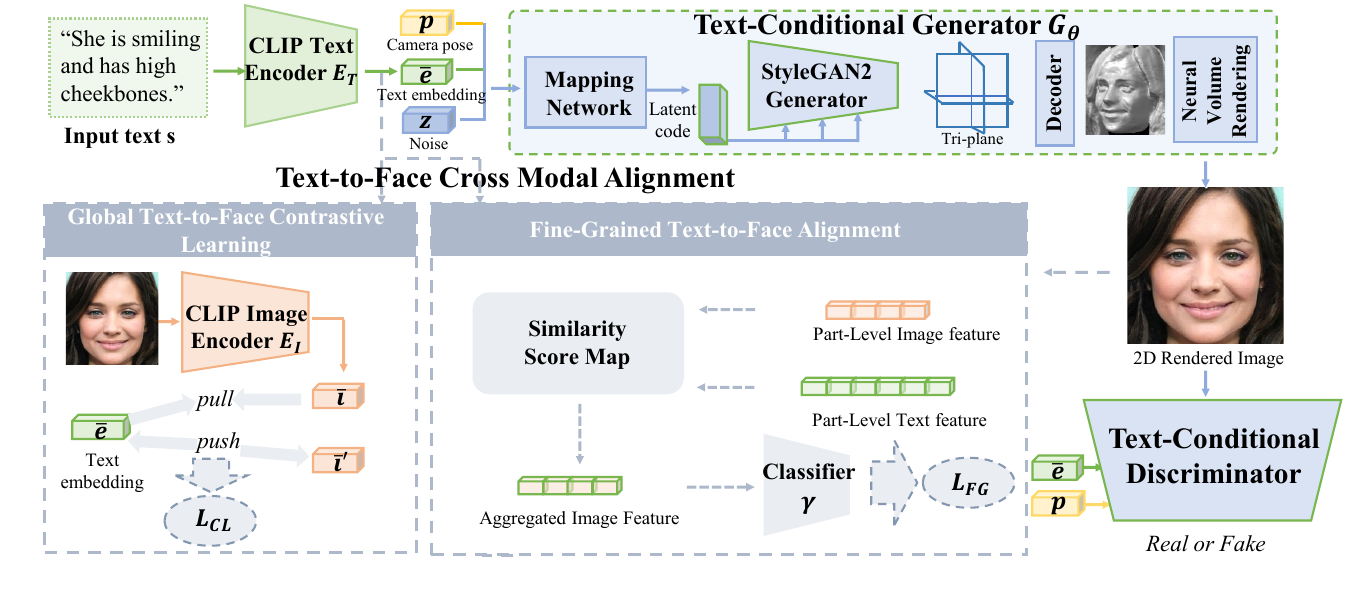}
\caption{
The framework of TG-3DFace. The text-conditional generator synthesizes 3D faces from text embedding $\bar{e}$ extracted by a pre-trained CLIP encoder, then renders them with camera parameters $p$. The text-conditional discriminator distinguishes real face images and the rendered fake face images and is trained adversarially against the generator. The text-to-face contrastive learning module provides a global text-face matching loss for the generator, enhancing the semantic consistency between the input texts and generated 3D shapes. The fine-grained text-to-face matching module helps the generator capture fine-grained semantic content of the input texts more accurately.
}
\vspace{-2mm}
\label{fig:framework}
\end{center}
\end{figure*}

\section{The Proposed Method}
\label{sec_method}
Our goal is to generate realistic 3D faces from text, facing the main challenges of limited text-3D face data and the requirement for semantic alignment between the generated faces and input texts.
To address these challenges, we propose TG-3DFace, as shown in Figure \ref{fig:framework}, which learns to generate 3D faces in condition of input text by using only text-2D face images. In this framework, global text-to-face contrastive learning and fine-grained text-to-face alignment are proposed to improve the semantic consistency between the generated 3D faces and the input texts.

\subsection{Text-conditional 3D Face Generation}
As illustrated in Figure \ref{fig:framework},
the text embedding $\bar{e}\in\mathbb{R}^{512}$ of input text $s$ is extracted by the CLIP text encoder $E_{T}$. The sentence embedding $\bar{e}$, camera parameters $p\in\mathbb{R}^{25}$ (including the intrinsic and extrinsic matrices), and a random noise $z\in\mathbb{R}^{512}$ are concatenated and projected into a latent code by the mapping network. This latent code then modulates the convolution kernels of the StyleGAN2 generator, producing a tri-plane representation of the 3D face from which 3D positions can be queried. The features of sampled 3D positions are aggregated and interpreted as a scalar density and a 32-channel feature by the decoder, both of which are then processed by a neural volume renderer \cite{rendering} to project the 3D feature volume into a 2D face image $\hat{x}$.
We define the mapping network, the StyleGAN2 generator, the decoder and the neural volume rendering as our text-conditional generator $G_{\theta}:\bar{e}\rightarrow \hat{x}$.
Subsequently, the text-conditional discriminator $D_{\phi}$ distinguishes real face images $x$ and the rendered fake images $\hat{x}$ based on the input text embedding $\bar{e}$ and camera parameters $p$.

\subsection{Text-to-Face Cross-Modal Alignment}
\subsubsection{Global Text-to-Face Contrastive Learning}
In order to generate 3D faces aligned with the input text, we propose global text-to-face contrastive learning to encourage the embeddings $(\bar{e},\bar{i})$ of paired text and face image close to each other, and unpaired ones $(\bar{e},\bar{i}^{'})$ away from each other,
where embeddings of text and images are extracted by the CLIP text encoder $E_{T}$ and CLIP image encoder $E_{I}$ respectively.
Global text-to-face contrastive learning encourages the generator to synthesize 3D faces that are semantically consistent with input text.
In particular, for each mini-batch, the image $\hat{x}_i$ generated from the input text $s_i$ is treated as a positive sample of $s_i$, while the other generated images $\hat{x}_j$ are regarded as negative samples. Positive and negative texts to the image $\hat{x_{i}}$ can be similarly defined.

Formally, the loss function for global text-to-face contrastive learning in a mini-batch is defined as:
\begin{equation}
   L_{CL} = \frac{1}{2n} \sum_{i=1}^n [L(\hat{x}_{i}) + L(s_{i})],
\end{equation}
where the loss for input text $s_i$ is defined as follows:
\begin{equation}
  L(s_{i})=- \frac{1}{n} \log\frac{\exp(E_{T}(s_i)\cdot E_{I}(\hat{x}_i)/\tau)}{\sum_{j=1}^n \exp(E_{T}(s_i)\cdot E_{I}(\hat{x}_j)/\tau)},
\end{equation}
where $E_{T}$ and $E_{I}$ are the CLIP text encoder and image encoder, $\tau$ is a temperature parameter, $n$ denotes the batch size, and the loss function $L(\hat{x}_{i})$ of contrastive learning for the generated image $\hat{x_{i}}$ can be similarly defined.

\subsubsection{Fine-grained Text-to-Face Alignment}
Until now, it is still challenging for the model to capture fine-grained facial attributes in the input text, as there is no fine-grained supervisions during training.
To this end, we explore fine-grained text-to-face alignment training signals for the text-conditional generator.

\begin{figure*}
\begin{center}
\includegraphics[width=1\linewidth]{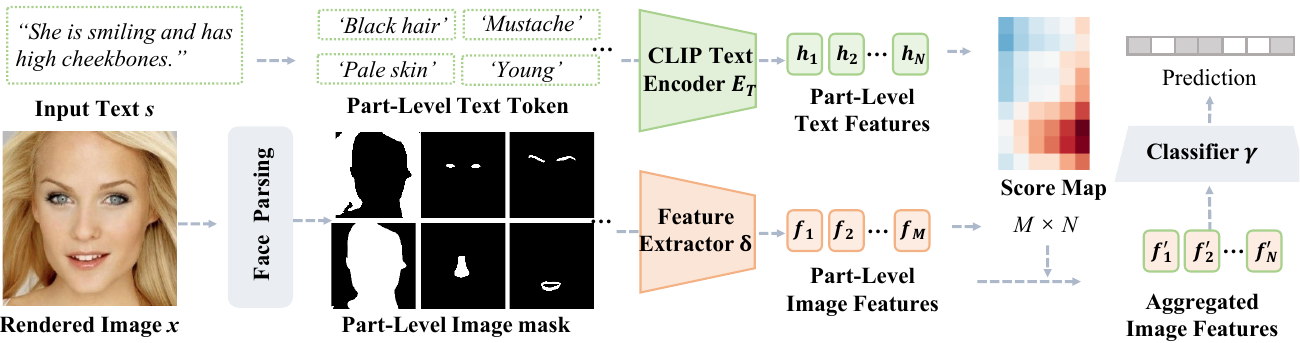}
\vspace{-2mm}
\end{center}
   \caption{Illustration of fine-grained text-to-face alignment. The rendered image is segmented into several part-level images by face parsing, and then part-level image features are extracted by a learned feature extractor $\delta$. The score map between part-level image features and part-level text features of pre-defined facial attributes text tokens can be used to aggregate the image features. Finally, the aggregated image features is used to predict facial attribute by passing through a learned classifier $\gamma$.}
\label{fig:short}
\end{figure*}

In fact, facial attributes are mainly displayed across several specific image areas. For example, the image area of the eyes corresponds to the facial attributes like ``blue eyes'', which means that different face regions have different contributions to an attribute.
Inspired by this, we propose to extract part-level image features, and align them with features of a set of pre-defined part-level texts about facial attributes, such as ``Black hair'', ``Mustache'', etc.
As the fine-grained text-to-face alignment module $C_{ \varphi}$ shown in Figure \ref{fig:short}, the rendered face image is first segmented into several parts by an off-the-shelf face parsing algorithm \cite{yu2018bisenet}.
Then, features of these part-level images are extracted by a feature extractor $\delta$, and a similarity matrix is established between these part-level image features and the fine-grained attribute text description. The part-level image features are then aggregated according to this similarity matrix, as the feature of the input face image. The aggregated feature is computed based on the feature similarity between the fine-grained text and part-level images, thereby focusing on the semantic information of the fine-grained attributes.

 Formally, the part-level image features $F = \{f_{i}\}_{i=1}^{M} \in\mathbb{R}^{M \times d}$ of an image are extracted by a part-level feature extractor $\delta:x_{p}\rightarrow \mathbb{R}^{512}$, where $x_{p}$ is a part-level face image in size of $512\times 512\times 3$.
 Part-level text features $H = \{h_{j}\}_{j=1}^{N} \in\mathbb{R}^{N\times d}$ of pre-defined part-level texts, such as ``black hair" and ``mustache" are extracted using the CLIP text encoder $E_{T}$, further projected to a matrix $K=l_K(H) \in\mathbb{R}^{N\times d}$ via learned linear projections $l_K$. The score maps are computed as:
\begin{equation}
    W = {\rm softmax}(\frac{FK^{T}}{\sqrt{d}}).
\end{equation}
The softmax function is applied to the rows of the scaled similarity matrix, where $d$ is the dimension of part-level image feature and text feature. The aggregated image feature $F^{'}\in\mathbb{R}^{N\times d}$ is then computed as $F^{'}=W^{T}F$.

Intuitively, the aggregated image feature $F^{'}$ is a weighted average of the part-level image features $F$, where the weights are the score map $W$, which are correlated to the similarity between $F$ and $K$.
The $i$-th row of $W$ is a normalized similarity vector between the part-level image feature $f_{i}$ and all the part-level token features $K$, such that more similar part-level image features have a higher contribution to the combination, whereas less similar part-level image features make little impact.
To further enhance the expressiveness of the fine-grained cross-modal alignment, multi-head attention \cite{vaswani2017attention} is utilized in parallel before passing the results through a learned classifier $\gamma$ for the final prediction of facial attributes of the input image.

Given the one-hot ground truth label $y\in\mathbb{R}^{k}$, the facial attribute classification loss can be calculated using the binary cross-entropy loss:
\begin{equation}
    L_{FG}=-\sum_{i=1}^{k}y^{(i)}\log(\gamma^{(i)}(F^{'})),
\end{equation}
where $y^{(i)}$ is the $i$-${th}$ ground truth label and $\gamma^{(i)}(F^{'})$ is the $i$-${th}$ predicted label of input image $x$ respectively. $k$ is the total number of selected attributes.

\subsection{Training Loss}
During training, the pretrained CLIP text encoder, image encoder, and the face parsing module are frozen,
The rest modules in the framework are trained end-to-end.
The text-conditional generator $G_{\theta}$, text-conditional discriminator $D_{\phi}$ and fine-grained text-to-face alignment module $C_{\varphi}$
play the following minimax game:
\begin{align}
     \min_{\theta,\varphi} \max_{\phi} &\frac{1}{n} \sum_{i=1}^n \{\log D_{\phi}(1 - G_{\theta}(z_i, s_i, p_i)) + \log D_{\phi}(x_i) \nonumber \\
     &+ [L_{FG}(G_{\theta}(z_i, s_i, p_i), y_i) + L_{FG}(x_i, y_i)]\} \nonumber\\
     &- \lVert \nabla D_{\phi}(x_i) \rVert^2 \nonumber\\
     &+ L_{CL},
 \end{align}
 where $z_i$, $s_i$, $p_i$, $x_i$ and $y_i$ are the $i$-th random noise, input text, camera parameter, real face image and attribute label in the mini-batch sampled from the training dataset,
 $n$ is the batch size.
 Specifically, we employ the non-saturating GAN loss function \cite{goodfellow2014gan}, where $G$ is trained to maximize $\log D_{\phi}(G_{\theta}(z_i, s_i, p_i))$ rather than $\log D_{\phi}(1 - G_{\theta}(z_i, s_i, p_i))$ to provide stronger gradients early in training.
The models are trained alternatively from scratch:  $D_{\phi}$ and $C_{\varphi}$ are firstly trained by one step, and then $G_{\theta}$ is trained for one step, until converges.

\subsection{Directional Classifier Guidance}
Generally, it is difficult to generate 3D faces from the out-of-domain input text such as
``He is a werewolf wearing glasses'' since that all our training data are photographs.
Inspired by classifier guidance in diffusion models \cite{nichol2021glide,ramesh2022dalle2,latent-diffusion},
that use an auxiliary discriminative model to guide the sampling process of a pretrained generative model,
we utilize the CLIP text encoder and image encoder to design the directional classifier guidance to guide the inference process so as to further improve the text-conditional generator towards generating out-of-domain 3D faces.

Given a target text like ``He is a werewolf wearing glasses" denoted as $s^{\star}$, our generator $G$ generates a 3D face firstly.
However, it may be a man wearing glasses as the generator $G$ has never seen werewolf. To perform directional classifier guidance,
we clone a copy $G_{frozen}$ of $G$ and freeze it afterwards, and then use $G_{frozen}$ and $G$ to generate a 3D face matching the target text $s^{\star}$.
The directional vector $V_{I_{i}}$ between the rendered face images from $G$ and $G_{frozen}$ in CLIP space can be obtained with the CLIP image encoder as
\begin{equation}
    V_{I_{i}} = E_{I}(G(z, s^{\star}, p_{i})) - E_{I}(G_{frozen}(z, s^{\star}, p_{i})).
\end{equation}
Similarly, we use a text prompt ``Photo" to describe style of training data,
noted as $s_o$, and then the directional vector $V_{T}$ between $s^{\star}$ and $s_{o}$ also can be obtained in CLIP space by the CLIP text encoder as
\begin{equation}
    V_{T} = E_{T}(s^{\star}) - E_{T}(s_{o}).
\end{equation}
We demand $V_{I_{i}}$ to be parallel to $V_{T}$, so that minimize the following directional classifier guidance loss:
\begin{equation}
    L_{DCG}=\frac{1}{M}\sum_{i=1}^M[1-\frac{V_{I_{i}}\cdot V_{T}}{|V_{I_{i}}||V_{T}|}],
\end{equation}
where $M$ is the number of randomly sampled camera poses in each optimization step.
Parameters of the text-conditional generator are updated by the directional classifier guidance loss to synthesize 3D faces matching text $s^{\star}$.
In experiments, we find that the changes in $s^{\star}$ can result in 3D faces with different styles.


\section{Experiments}
\label{sec_experiments}


\subsection{Datasets}
We conduct experiments on Multi-Modal CelebA-HQ \cite{xia2021tedigan} and CelebAText-HQ \cite{sea-t2f} datasets to verify the effectiveness of our method
for text-guided 3D face generation.
The Multi-Modal CelebA-HQ dataset has 30,000 face images, and each one has 10 text descriptions synthesized by facial attributes.
The CelebAText-HQ dataset contains 15,010 face images, in which each image has 10 manually annotated text descriptions.
All the face images come from the CelebA-HQ \cite{lee2020maskgan} dataset, in which each face image has an attribute annotation
related to 40 categories, such as ``Black hair'', ``Pale skin'', ``Young''.
These attributions are used as pre-defined part-level tokens in our method.
In order to learn better 3D face shapes, we added FFHQ \cite{stylegan}, a real-world human face dataset without corresponding text description, to the training set.
Off-the-shelf facial pose estimators \cite{Ekman1978FacialAC,schonberger2016structure} are used to extract approximate camera parameters for each face image in the training set.
The FFHQ-Text dataset \cite{FFHQ_text} is a face image dataset with large-scale facial attributes.
Since texts in the FFHQ-Text dataset are quite different from those in the Multi-Modal CelebA-HQ dataset, it can be used for cross-dataset experiment.


\subsection{Metrics}
We quantitatively evaluate the generated 3D faces in terms of the quality of their rendered 2D face images, including (1) the multi-view identity consistency (MVIC) by calculating the mean Arcface \cite{facenet} cosine similarity scores between pairs of face images of the same synthesized 3D face rendered from random camera poses;
(2) the reality and diversity of the rendered 2D face images, evaluated by the
Frechet Inception Distance (FID) \cite{heusel2017FID}; and
(3) the semantic consistency between the input texts and rendered 2D face images, measured by CLIP score.
The metric is defined as:
\begin{equation}
{\rm CLIP score}(x,s) = \max({\rm cosine}(E_{I}(x), E_{T}(s))\times 100, 0), \nonumber
\end{equation}
which corresponds to the cosine similarity between CLIP embeddings for an image $x$ and a text $s$ respectively in CLIP embedding space, $E_{I}$ and $E_{T}$ are CLIP image encoder and CLIP text encoder respectively.
Additional evaluation details are introduced in the supplemental materials.

\subsection{Main Results}
In this section, we first qualitatively verify the text-guided 3D face generation ability of our proposed TG-3DFace. Figure \ref{fig:main-result} shows the input texts and corresponding generated 3D faces, including rendered multi-view face images and 3D face meshes. As we can see, the multi-view face images are consistent with each other and the 3D meshes are detailed, indicating that TG-3DFace is able to generate high-quality 3D faces. Besides, fine-grained facial attributes including ``eyebrows", ``lipstick", ``hair", etc., can be well controlled by the input texts, indicating the text-to-face cross-modal alignment capability of TG-3DFace.

\subsection{Comparison on Text-to-3D Face Generation} 
\begin{figure}[t!]
\centering
\includegraphics[width=1\linewidth]{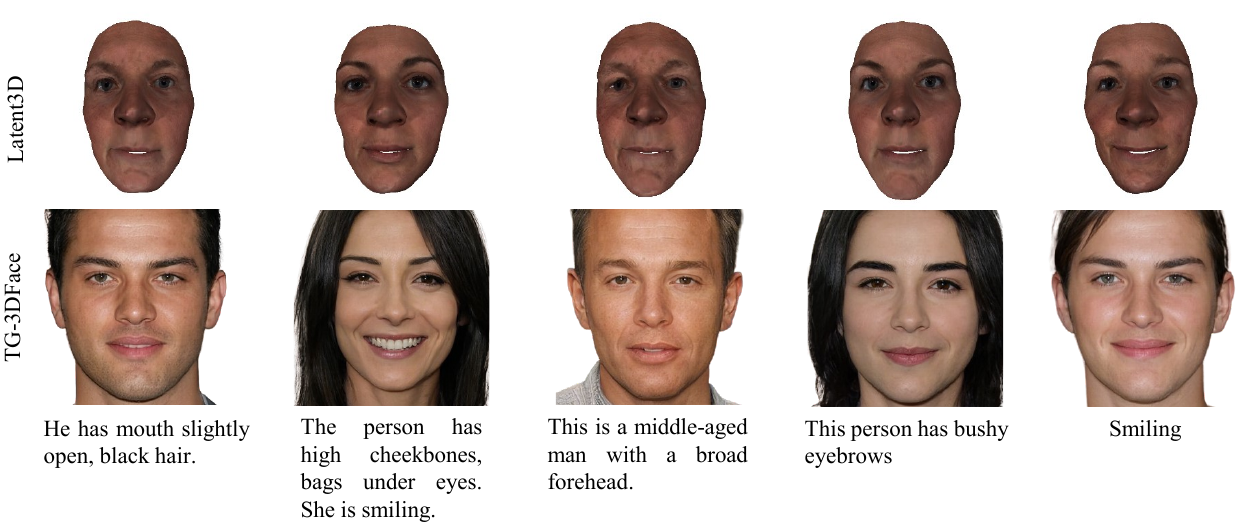}
\caption{Comparison on text-guided 3D Face generation.}
\label{fig:multi-view}
\end{figure}

\begin{figure}[t!]
\centering
\includegraphics[width=1\linewidth]{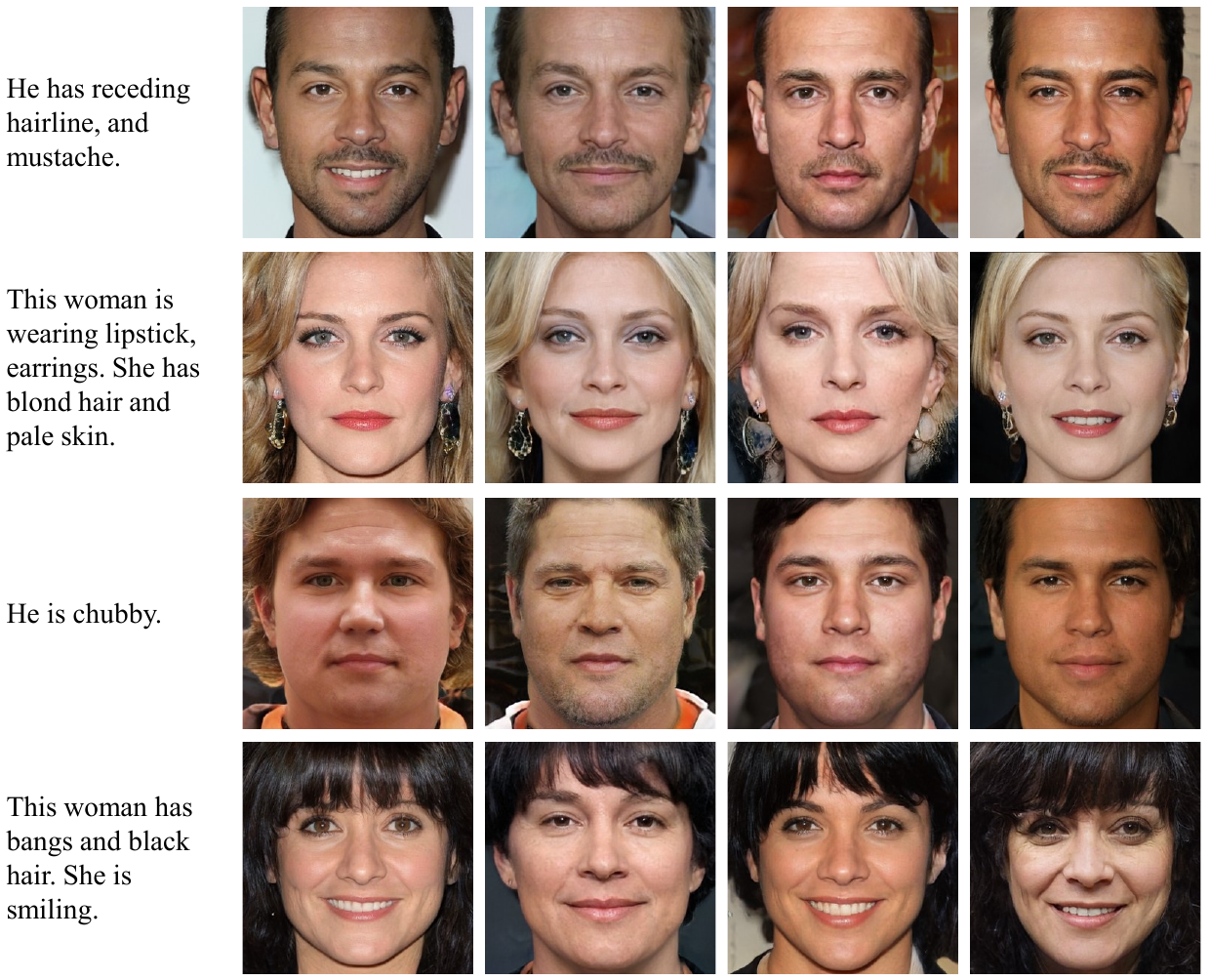}
\caption{Diverse generation results of TG-3DFace. In each row, we show input text and generated results of different input noises.}
\label{fig:diversity}
\end{figure}

We benchmark the text-guided 3D face generation ability of our proposed TG-3DFace by comparing against a text-to-3D face generation baseline method called Latent3D \cite{canfes2022text}.
Table \ref{tab:MVIC} shows the MVIC scores of Latent3D and TG-3DFace. As we can see, TG-3DFace achieves higher MVIC scores on both datasets,
which demonstrates that 3D faces generated by TG-3DFace have better multi-view consistency.
\begin{table}[b]
\centering
\begin{tabular}{lcc}
\hline
\multirow{2}{*}{Methods}                                                           & \multirow{2}{*}{\begin{tabular}[c]{@{}c@{}}Multi-Modal\\ CelebA-HQ\end{tabular}} & \multirow{2}{*}{\begin{tabular}[c]{@{}c@{}}CelebAText-\\ HQ\end{tabular}} \\
                                                                                   &                                                                                  &                                                                           \\
\hline
Latent3D \cite{canfes2022text} &0.87  &0.85  \\ \hline
\textbf{TG-3DFace}                                                                               &\textbf{0.95}                                                                            &\textbf{0.93}                                                      \\
\hline
\vspace{0.02in}
\end{tabular}
\caption{Quantitative comparisons on multi-view identity consistency (MVIC) against Latent3D \cite{canfes2022text} on the Multi-Modal CelebA-HQ and CelebAText-HQ datasets. MVIC is the higher the better.}
\label{tab:MVIC}
\end{table}
As shown in Figure \ref{fig:multi-view}, TG-3DFace can generate higher-quality 3D faces with detailed topology (e.g., hairs),
as well as realistic facial texture. The generated 3D faces of TG-3DFace with different input texts are also more diverse compared with Latent3D.
To further verify the diversity of our results, we show results from the same input text but different input noises in Figure \ref{fig:diversity}.
As we can see, given the same input text, TG-3DFace can generate diverse 3D faces according to different input noises.

\subsection{Comparison on Texture Quality}
In this section, we further benchmark the texture quality of generated 3D faces achieved by TG-3DFace and previous start-of-the-art
text-to-2D face/image generation methods, including SEA-T2F \cite{sea-t2f}, ControlGAN \cite{li2019controllable}, AttnGAN \cite{xu2018attngan} and AnyFace \cite{anyface}.
As shown in Table \ref{tab:multimodal} and Table \ref{tab:celebaTextHQ}, our method achieves better FID and CLIP score on both datasets, indicating the texture of generated 3D faces of TG-3DFace are of higher fidelity and semantic consistency.


\begin{table}[t!]
\begin{center}
\begin{tabular}{l|ccc}
\hline
Methods          & FID ↓  & CLIP Score ↑\\ \hline
SEA-T2F \cite{sea-t2f}          & 93.85   &20.81\\ \hline
ControlGAN \cite{li2019controllable}       & 74.59    &21.38\\ \hline
AttnGAN \cite{xu2018attngan}          & 51.69    &21.52\\ \hline
AnyFace \cite{anyface}          & 50.56    &-\\ \hline
\textbf{TG-3DFace}        & \bf{39.02}   &\bf{22.72}\\    \hline
\end{tabular}
\end{center}
\caption{Quantitative comparisons of different methods on the Multi-Modal CelebA-HQ dataset, where ↓ means the lower the better while ↑
means the opposite.}
\label{tab:multimodal}
\end{table}

\begin{table}[t!]
\begin{center}
\begin{tabular}{l|ccc}
\hline
Methods    & FID ↓   &CLIP Score ↑\\ \hline
SEA-T2F \cite{sea-t2f}    & 125.32     &19.06\\ \hline
ControlGAN \cite{li2019controllable} & 78.01      &20.70\\ \hline
AttnGAN \cite{xu2018attngan}    & 70.59      &20.17\\ \hline
AnyFace \cite{anyface}   & 56.75     &-\\ \hline
\textbf{TG-3DFace}  &\bf{52.21}  &\bf{21.03}\\ \hline
\end{tabular}
\end{center}
\caption{Quantitative comparisons of different methods on the CelebAText-HQ dataset, where ↓ means the lower the better while ↑
means the opposite.}
\label{tab:celebaTextHQ}
\end{table}

\begin{table}[t!]
\centering
\setlength{\tabcolsep}{1mm}{
\begin{tabular}{l|cc}
\hline
\multirow{2}{*}{Methods}                                                           & \multirow{2}{*}{\begin{tabular}[c]{@{}c@{}}Avg-rank on\\ Fidelity ↓ \end{tabular}} & \multirow{2}{*}{\begin{tabular}[c]{@{}c@{}}Avg-rank on\\ Semantic Consistency ↓ \end{tabular}} \\
                                                                                   &                                                                                  &                                                                           \\ \hline
SEA-T2F \cite{sea-t2f} &3.19  &4.00  \\ \hline
ControlGAN \cite{li2019controllable}                     &2.92  &2.61                                                                      \\ \hline
AttnGAN \cite{xu2018attngan}  &2.15  &2.11  \\ \hline
\textbf{TG-3DFace} &\textbf{1.02}  &\textbf{1.28}  \\ \hline
\end{tabular}}
\vspace{0.1in}
\caption{User study. Users show a significant preference for our TG-3DFace over SEA-T2F, ControlGAN and AttnGAN for fidelity and semantic consistency.}
\label{tab:user-study}
\end{table}

\begin{table}[t!]
\centering
\setlength{\tabcolsep}{1mm}{
\begin{tabular}{l|cc}
\hline
\multirow{2}{*}{Methods}                                                           & \multirow{2}{*}{\begin{tabular}[c]{@{}c@{}}Avg-rank on\\ Fidelity ↓ \end{tabular}} & \multirow{2}{*}{\begin{tabular}[c]{@{}c@{}}Avg-rank on\\ Semantic Consistency ↓ \end{tabular}} \\
                                                                                   &                                                                                  &                                                                           \\ \hline
SEA-T2F \cite{sea-t2f} &3.49  &3.18  \\ \hline
ControlGAN \cite{li2019controllable}                     &3.37  &3.16                                                                      \\ \hline
AttnGAN \cite{xu2018attngan}  &2.13  &2.36  \\ \hline
\textbf{TG-3DFace} &\textbf{1.00}  &\textbf{1.30}  \\ \hline

\end{tabular}}
\vspace{0.1in}
\caption{User study with out-of-distribution texts. Users show a significant preference for our TG-3DFace over SEA-T2F, ControlGAN and AttnGAN for fidelity and semantic consistency.}
\label{tab:user-study-ffhqtext}
\end{table}

\subsection{User Study}
We also employ a user study on the CelebAText-HQ dataset as an additional evaluation.
28 images generated from texts by different methods are randomly selected and 38 graduate students are invited to rank these images
with the questions ``\textit {Are these images real}" and ``\textit {Are these images achieve the attributes specified in the text}" (rank 1 is the best).
Ranking results for each method are averaged, defined as Avg-rank on Fidelity and Avg-rank on Semantic Consistency.
As shown in Table \ref{tab:user-study}, our model achieves better results, indicating our generated textures of 3D faces are of higher fidelity and semantic consistency with input texts.

To validate our TG-3DFace with out-of-distribution texts, we train our model and compared models on the Multi-Modal CelebA-HQ dataset, and test them using texts in the FFHQ-Text dataset.
We randomly select 28 images generated from texts by different models respectively
and invite 30 graduate students to rank them.
As listed in Table \ref{tab:user-study-ffhqtext}, our model achieves better results on user study with out-of-distribution texts.

\begin{figure}[t!]
\centering
\includegraphics[width=1\linewidth]{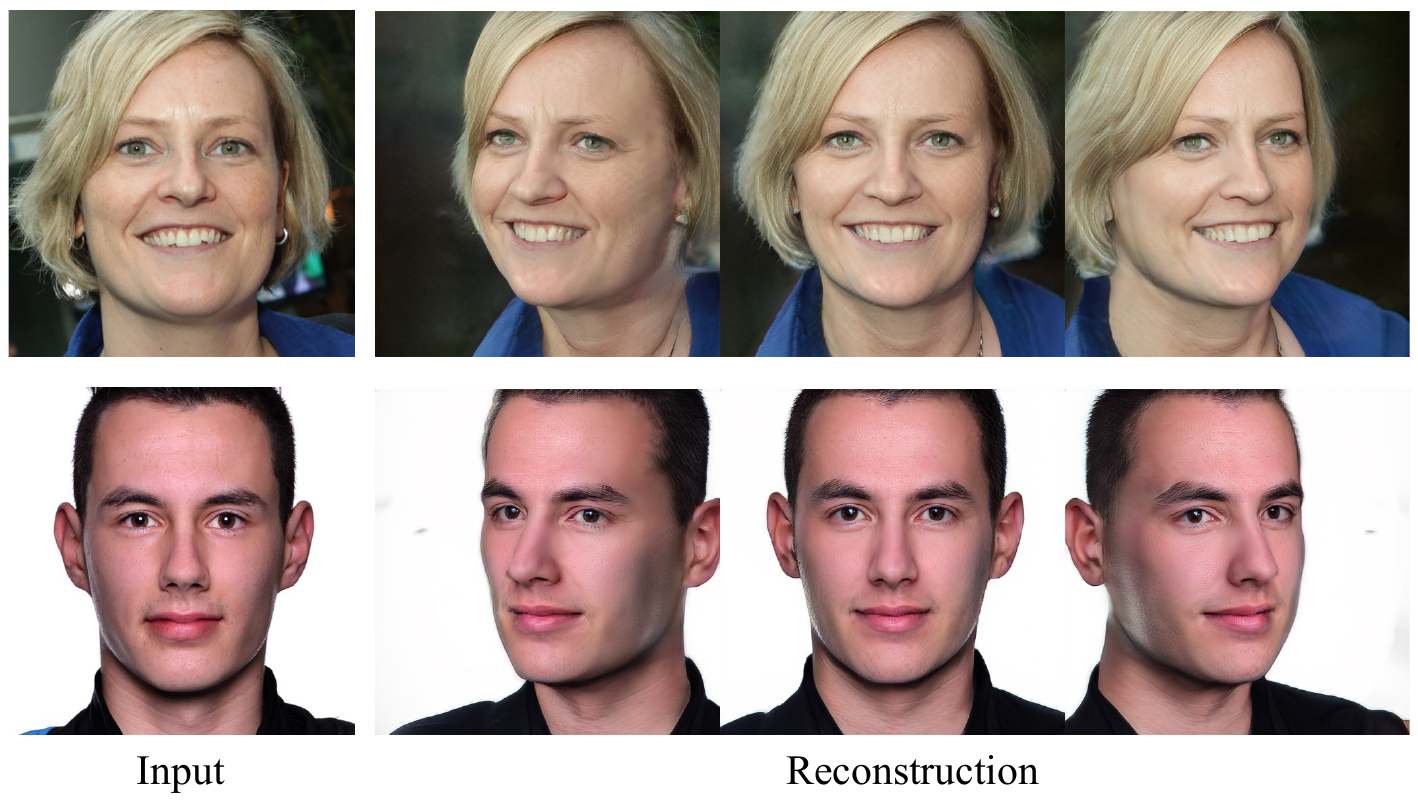}
\caption{Single-view 3D reconstruction results of TG-3DFace.}
\label{fig:inversion}
\end{figure}

\begin{table}[t]
\begin{center}
\begin{tabular}{l|ccc}
\hline
Methods          & FID ↓      & CLIP Score ↑\\ \hline
w/o $L_{CL}$    & 52.95     &21.50  &\\ \hline
w/o $L_{FG}$     &50.11	     &21.86\\ \hline
w/o $L_{FG}^{*}$ &52.57	      &22.03\\ \hline
Full model       &\bf{39.02}  &\bf{22.72}\\ \hline
\end{tabular}
\end{center}
\caption{Ablation study on Multi-Modal CelebA-HQ dataset, where ↓ means the lower the better, ↑ means the opposite.}
\label{tab:ablation_multi}
\end{table}

\begin{table}[t]
\begin{center}
\begin{tabular}{l|ccc}
\hline
Methods          & FID ↓ & CLIP Score ↑\\ \hline
w/o $L_{CL}$    &57.51  &20.61\\ \hline
w/o $L_{FG}$     &60.57	 &19.50\\ \hline
w/o $L_{FG}^{*}$ &56.75	 &20.73 \\ \hline
Full model       &\bf{52.21}  &\bf{21.03}\\ \hline
\end{tabular}
\end{center}
\caption{Ablation study on CelebAText-HQ dataset, where ↓ means the lower the better and ↑ means the higher the better.}
\label{tab:ablation_texthq}
\end{table}

\subsection{Ablation Studies}
To analyze the effectiveness of the proposed global text-to-face contrastive learning and fine-grained cross-modal alignment, we conduct ablation studies by removing one of them each time and report the quantitative results on the Multi-Modal CelebA-HQ dataset and CelebaTextHQ datasets.
As the results listed in Table \ref{tab:ablation_multi} and Table \ref{tab:ablation_texthq}, omitting $L_{CL}$ and $L_{FG}$ adversely affect the FID and CLIP score. Specifically, the FID increases from 39.02 to 52.95/50.11, and the CLIP score decreases from 22.72 to 21.50/21.86 on the Multi-Modal-celebA-HQ dataset.
This highlights the importance of
these two modules
in improving the quality of generated 3D faces and enhancing the semantic matching between the generated 3D faces and the input texts.

To further emphasize the significance of the fine-grained text-to-face alignment module,
we retain its network and loss while removing the operations of face parsing and part-level feature aggregation, and extract features from the full image to predict its attributes instead, denoted as ``w/o $L_{FG}^{*}$''. The comparison between ``w/o $L_{FG}^{*}$" and the full model demonstrates the importance of our carefully designed fine-grained cross-modal alignment, ruling out the possibility of it being replaced by an attribute classifier on the full image.




\begin{figure}[t]
\centering
\includegraphics[width=1\linewidth]{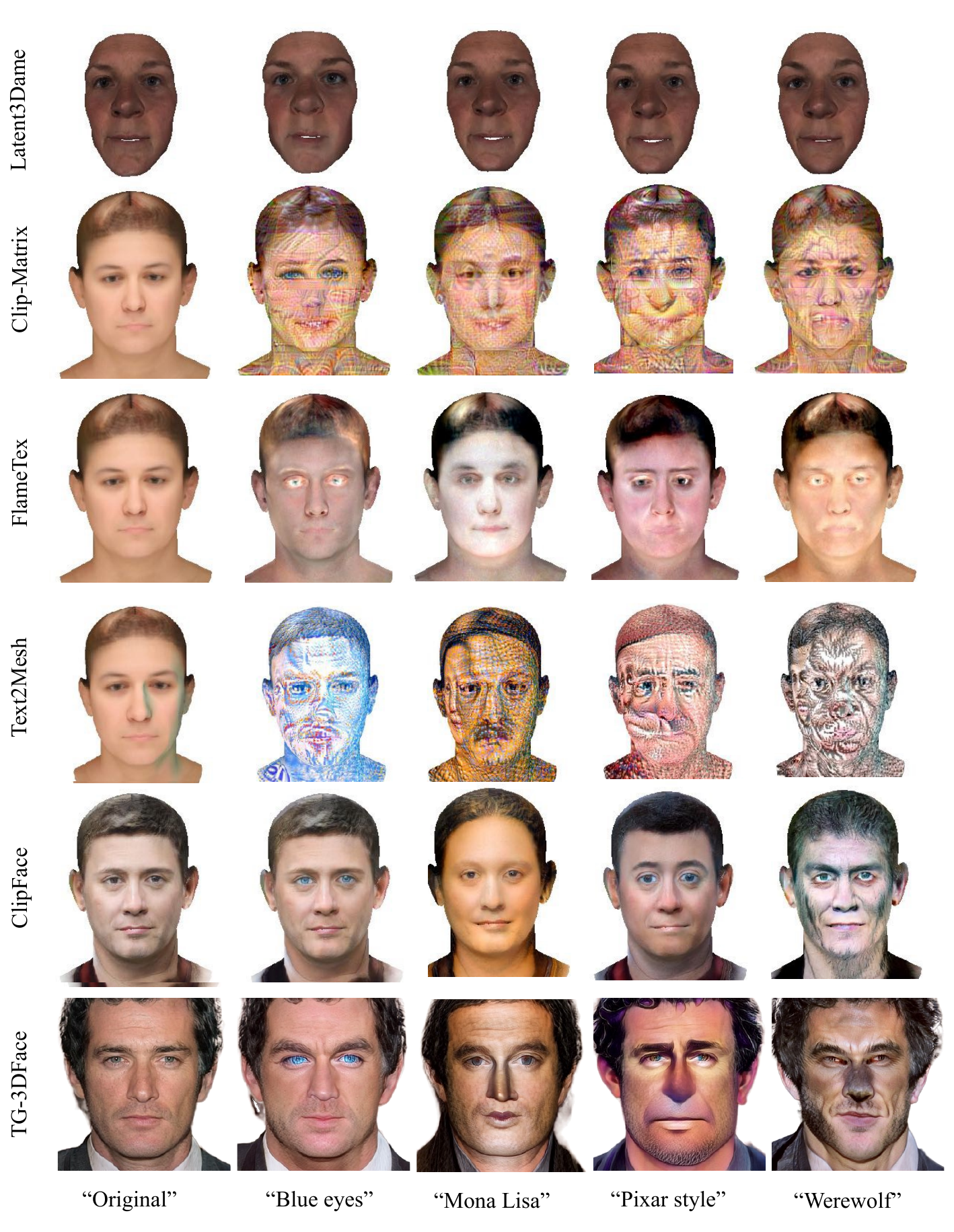}
\vspace{-0.15in}
\caption{Comparison on text-guided 3D face manipulation.}
\label{fig:edit}
\end{figure}

\subsection{Applications}
\textbf{Single-view 3D Face Reconstruction.}
Figure \ref{fig:inversion} illustrates the utilization of our learned latent space for single-view 3D reconstruction via pivotal tuning inversion (PTI) \cite{roich2022pivotal}. The 3D prior over text-guided 3D face generation enables impressive single-view geometry recovery.
Building on this ability,
we can further edit the reconstruction result using directional classifier guidance, which may serve as a promising area for future research.

\textbf{Text-guided 3D Face Manipulation.}
The proposed directional classifier guidance enables our method to create many interesting results in diverse styles based on our learned text-guided 3D face generative model.
Given a text that differs significantly from the texts in training data during inference time, the proposed directional classifier guidance can be used to optimize the generator for a few minutes to synthesize 3D faces with styles outside the training set.
Figure \ref{fig:edit} compares 3D face editing results achieved by different text-driven 3D texture manipulation methods, including Latent3D \cite{canfes2022text}, Clip-Matrix \cite{jetchev2021clipmatrix}, FlameTex$^1$,
Text2Mesh \cite{text2mesh}, ClipFace \cite{aneja2023clipface} and TG-3DFace.
As we can see,
Clip-Matrix, FlameTex, and Text2Mesh
cannot get the correct texture according to the text.
Latent3D cannot capture finer-grained localization of manipulations, such as changing the color of eyes, and it cannot generate textures in various styles, such as Pixar.
Although ClipFace can generate the corresponding texture of 3D faces according to the input texts,
they can not handle accessories like
headwear, or eyewear, due to the use of the FLAME \cite{flame} model, which does not capture accessories
or complex hair.
Our approach yields consistently
high-quality textures for various prompts, in comparison to these
baselines.
In general, our generator enables high-quality editing, and the style will be more obvious in texture.
\footnotetext{$^1$ https://github.com/HavenFeng/photometric\_optimization}

\subsection{Parameters and Runtime}
Table \ref{tab:para} compares parameters and inference time running on a single NVIDIA Tesla V100 GPU between our TG-3DFace and several existing text-guided 3D face or object generation methods.
We can see that when the model parameters of TG-3DFace are not large, the inference time to generate a 3D face is only 0.05 seconds, and the manipulation time is only 1.5 minutes.
\begin{table}[h!]
\centering
\setlength{\tabcolsep}{1mm}{
\begin{tabular}{l|ccc}
\hline
\multirow{2}{*}{Methods}                                                           & \multirow{2}{*}{\begin{tabular}[c]{@{}c@{}}Total \\Params \end{tabular}} & \multirow{2}{*}{\begin{tabular}[c]{@{}c@{}}Trainable \\ Params \end{tabular}} &
\multirow{2}{*}{\begin{tabular}[c]{@{}c@{}}Inference\\ Time \end{tabular}} \\
 &               &                        \\ \hline
Latent3D \cite{canfes2022text}    &661 M     &-    &6 min\\ \hline
Clip-Matrix \cite{jetchev2021clipmatrix} &154 M &2.6 M    &30 min  \\ \hline
Text2Mesh \cite{text2mesh}   &151 M & 659 K   &25 min\\ \hline
TG-3DFace (generation) &240 M  &76 M   &0.05 s\\    \hline
TG-3DFace (manipulation) &190 M & 39 M   &1.5 min\\   \hline
\end{tabular}
}
\vspace{0.1in}
\caption{Comparison of total parameters, trainable paramateres, and inference time per sample.}
\label{tab:para}
\end{table}

\section{Implementation Details}
We train our model with a batch size of 32, and
use a discriminator learning rate of 0.002 and a generator learning rate of 0.0025.
Similar to EG3D \cite{eg3d}, we blur images when they enter the discriminator, gradually reducing the amount of blur of the first 200 K images, and we train our model without style mixing regularization.
According to EG3D, low neural rendering resolutions (e.g., 64) enable faster speed of training and inference, while higher neural rendering resolutions (e.g., 128) facilitate more detailed shapes and more view-consistent 3D renderings.
Following EG3D, the neural rendering resolution is gradually increased from 64$^2$ to 128$^2$ over 1 million images during training. The total training time of our model on 8 NVIDIA Tesla A100 GPUs is 48 hours.
When the directional classifier guidance is used in the inference phase, the generator is optimized for 100 iterations at a learning rate of 0.002. 


\section{Conclusion}
\label{sec_conclusion}
In this paper, we propose a novel method named TG-3DFace for generating high-quality 3D faces with multi-view consistent and photo-realistic rendered face images.
Specifically, a text-conditional 3D face GAN enables the model can be trained from text-face images rather than the supervision of 3D faces.
Global text-to-face contrastive learning and fine-grained text-to-face alignment modules are proposed to improve the semantic consistency between the generated 3D faces and input texts.
Furthermore, we extend our model to synthesize out-of-domain 3D faces by introducing directional classifier guidance.
Extensive experimental studies manifest the effectiveness of our method.

\section{Limitation and Future Work}
First, our method cannot infer identity information from textual descriptions, such as ``Toms Bond".
Second, the 3D faces generated by our method are sometimes asymmetry, such as wearing only one earring.
Third, the race of the generated faces is similar to that of training data.
We will consider improving the quality of generated shape,
and will expand the races of training images so that the resulting faces are not limited to a single race.

\section{Acknowledgement}
The work was supported in part by the National Natural Science Foundation of China (NSFC) under Grant (No. 61976173, 12125104, U20B2075), Shaanxi Fundamental Science Research Project for Mathematics and Physics (Grant No. 22JSY011). We thank MindSpore for the partial support of this work, which is a new deep learning computing framework$^2$.
\footnotetext{$^2$ https://www.mindspore.cn}


\begin{thebibliography}{100}\itemsep=-1pt

\bibitem{aneja2023clipface}
Shivangi Aneja, Justus Thies, Angela Dai, and Matthias Nie{\ss}ner.
\newblock Clip{F}ace: {T}ext-guided editing of textured 3{D} morphable models.
\newblock In {\em SIGGRAPH '23 Conference Proceedings}, 2023.

\bibitem{arjovsky2017wasserstein}
Martin Arjovsky, Soumith Chintala, and L{\'e}on Bottou.
\newblock Wasserstein generative adversarial networks.
\newblock In {\em International conference on machine learning}, pages
  214--223, 2017.

\bibitem{athar2022rignerf}
ShahRukh Athar, Zexiang Xu, Kalyan Sunkavalli, Eli Shechtman, and Zhixin Shu.
\newblock Rig{N}e{RF}: {F}ully controllable neural 3{D} portraits.
\newblock In {\em Proceedings of the IEEE/CVF Conference on Computer Vision and
  Pattern Recognition}, pages 20364--20373, 2022.

\bibitem{bengio2013estimating}
Yoshua Bengio, Nicholas L{\'e}onard, and Aaron Courville.
\newblock Estimating or propagating gradients through stochastic neurons for
  conditional computation.
\newblock {\em arXiv preprint arXiv:1308.3432}, 2013.

\bibitem{kid_score}
Miko{\l}aj Bi{\'n}kowski, Dougal~J Sutherland, Michael Arbel, and Arthur
  Gretton.
\newblock Demystifying {MMD} {GAN}s.
\newblock {\em arXiv preprint arXiv:1801.01401}, 2018.

\bibitem{canfes2022text}
Zehranaz Canfes, M~Furkan Atasoy, Alara Dirik, and Pinar Yanardag.
\newblock Text and image guided 3{D} avatar generation and manipulation.
\newblock {\em arXiv preprint arXiv:2202.06079}, 2022.

\bibitem{cao2017colorgan}
Yun Cao, Zhiming Zhou, Weinan Zhang, and Yong Yu.
\newblock Unsupervised diverse colorization via generative adversarial
  networks.
\newblock In {\em Joint European conference on machine learning and knowledge
  discovery in databases}, pages 151--166, 2017.

\bibitem{eg3d}
Eric~R Chan, Connor~Z Lin, Matthew~A Chan, Koki Nagano, Boxiao Pan, Shalini
  De~Mello, Orazio Gallo, Leonidas~J Guibas, Jonathan Tremblay, Sameh Khamis,
  et~al.
\newblock Efficient geometry-aware 3{D} generative adversarial networks.
\newblock In {\em Proceedings of the IEEE/CVF Conference on Computer Vision and
  Pattern Recognition}, pages 16123--16133, 2022.

\bibitem{pigan}
Eric~R Chan, Marco Monteiro, Petr Kellnhofer, Jiajun Wu, and Gordon Wetzstein.
\newblock Pi-{GAN}: {P}eriodic implicit generative adversarial networks for
  3{D}-aware image synthesis.
\newblock In {\em Proceedings of the IEEE/CVF Conference on Computer Vision and
  Pattern Recognition}, pages 5799--5809, 2021.

\bibitem{chang2015shapenet}
Angel~X Chang, Thomas Funkhouser, Leonidas Guibas, Pat Hanrahan, Qixing Huang,
  Zimo Li, Silvio Savarese, Manolis Savva, Shuran Song, Hao Su, et~al.
\newblock Shape{N}et: {A}n information-rich 3{D} model repository.
\newblock {\em arXiv preprint arXiv:1512.03012}, 2015.

\bibitem{changpinyo2021cc12m}
Soravit Changpinyo, Piyush Sharma, Nan Ding, and Radu Soricut.
\newblock Conceptual 12{M}: {P}ushing web-scale image-text pre-training to
  recognize long-tail visual concepts.
\newblock In {\em Proceedings of the IEEE/CVF Conference on Computer Vision and
  Pattern Recognition}, pages 3558--3568, 2021.

\bibitem{chen2018language}
Jianbo Chen, Yelong Shen, Jianfeng Gao, Jingjing Liu, and Xiaodong Liu.
\newblock Language-based image editing with recurrent attentive models.
\newblock In {\em Proceedings of the IEEE/CVF Conference on Computer Vision and
  Pattern Recognition}, pages 8721--8729, 2018.

\bibitem{chen2018text2shape}
Kevin Chen, Christopher~B Choy, Manolis Savva, Angel~X Chang, Thomas
  Funkhouser, and Silvio Savarese.
\newblock Text2shape: Generating shapes from natural language by learning joint
  embeddings.
\newblock In {\em Asian conference on computer vision}, pages 100--116, 2018.

\bibitem{chen2017cascaded_refinement}
Qifeng Chen and Vladlen Koltun.
\newblock Photographic image synthesis with cascaded refinement networks.
\newblock In {\em Proceedings of the IEEE international conference on computer
  vision}, pages 1511--1520, 2017.

\bibitem{chen2020uniter}
Yen-Chun Chen, Linjie Li, Licheng Yu, Ahmed El~Kholy, Faisal Ahmed, Zhe Gan, Yu
  Cheng, and Jingjing Liu.
\newblock Uniter: Universal image-text representation learning.
\newblock In {\em European conference on computer vision}, pages 104--120,
  2020.

\bibitem{collins2021abo}
Jasmine Collins, Shubham Goel, Kenan Deng, Achleshwar Luthra, Leon Xu, Erhan
  Gundogdu, Xi Zhang, Tomas F~Yago Vicente, Thomas Dideriksen, Himanshu Arora,
  et~al.
\newblock Abo: Dataset and benchmarks for real-world 3d object understanding.
\newblock In {\em Proceedings of the IEEE/CVF Conference on Computer Vision and
  Pattern Recognition}, pages 21126--21136, 2022.

\bibitem{blender2008}
Blender~Online Community.
\newblock Blender - a 3d modelling and rendering package.
\newblock {\em Blender Foundation, Stichting Blender Foundation, Amsterdam},
  2018.

\bibitem{dauphin2017language}
Yann~N Dauphin, Angela Fan, Michael Auli, and David Grangier.
\newblock Language modeling with gated convolutional networks.
\newblock In {\em International conference on machine learning}, pages
  933--941, 2017.

\bibitem{deng2019arcface}
Jiankang Deng, Jia Guo, Niannan Xue, and Stefanos Zafeiriou.
\newblock Arcface: {A}dditive angular margin loss for deep face recognition.
\newblock In {\em Proceedings of the IEEE/CVF conference on computer vision and
  pattern recognition}, pages 4690--4699, 2019.

\bibitem{devlin2018bert}
Jacob Devlin, Ming-Wei Chang, Kenton Lee, and Kristina Toutanova.
\newblock {BERT}: {P}re-training of deep bidirectional transformers for
  language understanding.
\newblock {\em arXiv preprint arXiv:1810.04805}, 2018.

\bibitem{dhariwal2021adm}
Prafulla Dhariwal and Alexander Nichol.
\newblock Diffusion models beat gans on image synthesis.
\newblock {\em Advances in Neural Information Processing Systems}, 34, 2021.

\bibitem{ding2021cogview}
Ming Ding, Zhuoyi Yang, Wenyi Hong, Wendi Zheng, Chang Zhou, Da Yin, Junyang
  Lin, Xu Zou, Zhou Shao, Hongxia Yang, et~al.
\newblock Cog{V}iew: {M}astering text-to-image generation via transformers.
\newblock {\em Advances in Neural Information Processing Systems},
  34:19822--19835, 2021.

\bibitem{ding2022cogview2}
Ming Ding, Wendi Zheng, Wenyi Hong, and Jie Tang.
\newblock Cog{V}iew2: Faster and better text-to-image generation via
  hierarchical transformers.
\newblock {\em arXiv preprint arXiv:2204.14217}, 2022.

\bibitem{sisgan}
Hao Dong, Simiao Yu, Chao Wu, and Yike Guo.
\newblock Semantic image synthesis via adversarial learning.
\newblock In {\em Proceedings of the IEEE International Conference on Computer
  Vision}, pages 5706--5714, 2017.

\bibitem{dosovitskiy2020vit}
Alexey Dosovitskiy, Lucas Beyer, Alexander Kolesnikov, Dirk Weissenborn,
  Xiaohua Zhai, Thomas Unterthiner, Mostafa Dehghani, Matthias Minderer, Georg
  Heigold, Sylvain Gelly, et~al.
\newblock An image is worth 16x16 words: Transformers for image recognition at
  scale.
\newblock In {\em International Conference on Learning Representations}, 2020.

\bibitem{Ekman1978FacialAC}
Paul Ekman and Wallace~V. Friesen.
\newblock Facial action coding system: {A} technique for the measurement of
  facial movement.
\newblock 1978.

\bibitem{el2019tell}
Alaaeldin El-Nouby, Shikhar Sharma, Hannes Schulz, Devon Hjelm, Layla~El Asri,
  Samira~Ebrahimi Kahou, Yoshua Bengio, and Graham~W Taylor.
\newblock Tell, draw, and repeat: Generating and modifying images based on
  continual linguistic instruction.
\newblock In {\em Proceedings of the IEEE/CVF International Conference on
  Computer Vision}, pages 10304--10312, 2019.

\bibitem{esser2021imagebart}
Patrick Esser, Robin Rombach, Andreas Blattmann, and Bjorn Ommer.
\newblock Image{BART}: {B}idirectional context with multinomial diffusion for
  autoregressive image synthesis.
\newblock {\em Advances in Neural Information Processing Systems},
  34:3518--3532, 2021.

\bibitem{vqgan}
Patrick Esser, Robin Rombach, and Bjorn Ommer.
\newblock Taming transformers for high-resolution image synthesis.
\newblock In {\em Proceedings of the IEEE/CVF Conference on Computer Vision and
  Pattern Recognition}, pages 12873--12883, 2021.

\bibitem{deca}
Yao Feng, Haiwen Feng, Michael~J Black, and Timo Bolkart.
\newblock Learning an animatable detailed 3{D} face model from in-the-wild
  images.
\newblock {\em ACM Transactions on Graphics}, 40(4):1--13, 2021.

\bibitem{fu2022shapecrafter}
Rao Fu, Xiao Zhan, Yiwen Chen, Daniel Ritchie, and Srinath Sridhar.
\newblock Shape{C}rafter: {A} recursive text-conditioned 3{D} shape generation
  model.
\newblock {\em arXiv preprint arXiv:2207.09446}, 2022.

\bibitem{fu2020iterative}
Tsu-Jui Fu, Xin Wang, Scott Grafton, Miguel Eckstein, and William~Yang Wang.
\newblock Iterative language-based image editing via self-supervised
  counterfactual reasoning.
\newblock In {\em Proceedings of the 2020 Conference on Empirical Methods in
  Natural Language Processing (EMNLP)}, pages 4413--4422, 2020.

\bibitem{gal2022stylegan}
Rinon Gal, Or Patashnik, Haggai Maron, Amit~H Bermano, Gal Chechik, and Daniel
  Cohen-Or.
\newblock Style{GAN-NADA}: {C}lip-guided domain adaptation of image generators.
\newblock {\em ACM Transactions on Graphics}, 41(4):1--13, 2022.

\bibitem{style_transfer}
Leon~A Gatys, Alexander~S Ecker, and Matthias Bethge.
\newblock Image style transfer using convolutional neural networks.
\newblock In {\em Proceedings of the IEEE/CVF conference on computer vision and
  pattern recognition}, pages 2414--2423, 2016.

\bibitem{fastrcnn}
Ross Girshick.
\newblock Fast r-cnn.
\newblock In {\em Proceedings of the IEEE international conference on computer
  vision}, pages 1440--1448, 2015.

\bibitem{rcnn}
Ross Girshick, Jeff Donahue, Trevor Darrell, and Jitendra Malik.
\newblock Rich feature hierarchies for accurate object detection and semantic
  segmentation.
\newblock In {\em Proceedings of the IEEE/CVF conference on computer vision and
  pattern recognition}, pages 580--587, 2014.

\bibitem{goodfellow2014gan}
Ian Goodfellow, Jean Pouget-Abadie, Mehdi Mirza, Bing Xu, David Warde-Farley,
  Sherjil Ozair, Aaron Courville, and Yoshua Bengio.
\newblock Generative adversarial nets.
\newblock {\em Advances in Neural Information Processing Systems}, 27, 2014.

\bibitem{instance_segmentation}
Kaiming He, Georgia Gkioxari, Piotr Doll{\'a}r, and Ross Girshick.
\newblock Mask {R-CNN}.
\newblock In {\em Proceedings of the IEEE international conference on computer
  vision}, pages 2961--2969, 2017.

\bibitem{resnet}
Kaiming He, Xiangyu Zhang, Shaoqing Ren, and Jian Sun.
\newblock Deep residual learning for image recognition.
\newblock In {\em Proceedings of the IEEE/CVF conference on computer vision and
  pattern recognition}, pages 770--778, 2016.

\bibitem{hendrycks2016gaussian}
Dan Hendrycks and Kevin Gimpel.
\newblock Gaussian error linear units (gelus).
\newblock {\em arXiv preprint arXiv:1606.08415}, 2016.

\bibitem{heusel2017FID}
Martin Heusel, Hubert Ramsauer, Thomas Unterthiner, Bernhard Nessler, and Sepp
  Hochreiter.
\newblock {GAN}s trained by a two time-scale update rule converge to a local
  nash equilibrium.
\newblock {\em Advances in Neural Information Processing Systems}, 30, 2017.

\bibitem{ho2020ddpm}
Jonathan Ho, Ajay Jain, and Pieter Abbeel.
\newblock Denoising diffusion probabilistic models.
\newblock {\em Advances in Neural Information Processing Systems},
  33:6840--6851, 2020.

\bibitem{ho2019axial}
Jonathan Ho, Nal Kalchbrenner, Dirk Weissenborn, and Tim Salimans.
\newblock Axial attention in multidimensional transformers.
\newblock {\em arXiv preprint arXiv:1912.12180}, 2019.

\bibitem{ho2021classifier-free-guidance}
Jonathan Ho and Tim Salimans.
\newblock Classifier-free diffusion guidance.
\newblock In {\em NeurIPS 2021 Workshop on Deep Generative Models and
  Downstream Applications}, 2021.

\bibitem{hong2022avatarclip}
Fangzhou Hong, Mingyuan Zhang, Liang Pan, Zhongang Cai, Lei Yang, and Ziwei
  Liu.
\newblock Avatar{CLIP}: {Z}ero-shot text-driven generation and animation of
  3{D} avatars.
\newblock {\em ACM Transactions on Graphics}, 41(4):1--19, 2022.

\bibitem{hu2021text}
Li Hu, Jinwei Qi, Bang Zhang, Pan Pan, and Yinghui Xu.
\newblock Text-driven 3{D} avatar animation with emotional and expressive
  behaviors.
\newblock In {\em Proceedings of the ACM International Conference on
  Multimedia}, pages 2816--2818, 2021.

\bibitem{isola2017pix2pix}
Phillip Isola, Jun-Yan Zhu, Tinghui Zhou, and Alexei~A Efros.
\newblock Image-to-image translation with conditional adversarial networks.
\newblock In {\em Proceedings of the IEEE/CVF conference on computer vision and
  pattern recognition}, pages 1125--1134, 2017.

\bibitem{jahan2021semantics}
Tansin Jahan, Yanran Guan, and Oliver van Kaick.
\newblock Semantics-guided latent space exploration for shape generation.
\newblock In {\em Computer Graphics Forum}, volume~40, pages 115--126, 2021.

\bibitem{jain2021dreamfields}
Ajay Jain, Ben Mildenhall, Jonathan~T Barron, Pieter Abbeel, and Ben Poole.
\newblock Zero-shot text-guided object generation with dream fields.
\newblock In {\em Proceedings of the IEEE/CVF Conference on Computer Vision and
  Pattern Recognition}, pages 867--876, 2022.

\bibitem{jetchev2021clipmatrix}
Nikolay Jetchev.
\newblock Clip{M}atrix: {T}ext-controlled creation of 3{D} textured meshes.
\newblock {\em arXiv preprint arXiv:2109.12922}, 2021.

\bibitem{jia2021align}
Chao Jia, Yinfei Yang, Ye Xia, Yi-Ting Chen, Zarana Parekh, Hieu Pham, Quoc Le,
  Yun-Hsuan Sung, Zhen Li, and Tom Duerig.
\newblock Scaling up visual and vision-language representation learning with
  noisy text supervision.
\newblock In {\em International Conference on Machine Learning}, pages
  4904--4916, 2021.

\bibitem{jiang2021language}
Wentao Jiang, Ning Xu, Jiayun Wang, Chen Gao, Jing Shi, Zhe Lin, and Si Liu.
\newblock Language-guided global image editing via cross-modal cyclic
  mechanism.
\newblock In {\em Proceedings of the IEEE/CVF International Conference on
  Computer Vision}, pages 2115--2124, 2021.

\bibitem{johnson2016perceptual}
Justin Johnson, Alexandre Alahi, and Li Fei-Fei.
\newblock Perceptual losses for real-time style transfer and super-resolution.
\newblock In {\em European conference on computer vision}, pages 694--711,
  2016.

\bibitem{stylegan}
Tero Karras, Samuli Laine, and Timo Aila.
\newblock A style-based generator architecture for generative adversarial
  networks.
\newblock In {\em Proceedings of the IEEE/CVF Conference on Computer Vision and
  Pattern Recognition}, pages 4401--4410, 2019.

\bibitem{karras2019style}
Tero Karras, Samuli Laine, and Timo Aila.
\newblock A style-based generator architecture for generative adversarial
  networks.
\newblock In {\em Proceedings of the IEEE/CVF Conference on Computer Vision and
  Pattern Recognition}, pages 4401--4410, 2019.

\bibitem{stylegan2}
Tero Karras, Samuli Laine, Miika Aittala, Janne Hellsten, Jaakko Lehtinen, and
  Timo Aila.
\newblock Analyzing and improving the image quality of {S}tyle{GAN}.
\newblock In {\em Proceedings of the IEEE/CVF Conference on Computer Vision and
  Pattern Recognition}, pages 8110--8119, 2020.

\bibitem{khalid2022clip-mesh}
Nasir Khalid, Tianhao Xie, Eugene Belilovsky, and Tiberiu Popa.
\newblock {CLIP}-{M}esh: {G}enerating textured meshes from text using
  pretrained image-text models.
\newblock {\em ACM Transactions on Graphics}, 2022.

\bibitem{khalid2022text}
Nasir Khalid, Tianhao Xie, Eugene Belilovsky, and Tiberiu Popa.
\newblock Text to mesh without 3{D} supervision using limit subdivision.
\newblock {\em arXiv preprint arXiv:2203.13333}, 2022.

\bibitem{kim2017discogan}
Taeksoo Kim, Moonsu Cha, Hyunsoo Kim, Jung~Kwon Lee, and Jiwon Kim.
\newblock Learning to discover cross-domain relations with generative
  adversarial networks.
\newblock In {\em International Conference on Machine Learning}, pages
  1857--1865, 2017.

\bibitem{2014Auto}
D.~P. Kingma and M. Welling.
\newblock Auto-encoding variational bayes.
\newblock {\em ICLR}, 2014.

\bibitem{panoptic_segmentation}
Alexander Kirillov, Kaiming He, Ross Girshick, Carsten Rother, and Piotr
  Doll{\'a}r.
\newblock Panoptic segmentation.
\newblock In {\em Proceedings of the IEEE/CVF Conference on Computer Vision and
  Pattern Recognition}, pages 9404--9413, 2019.

\bibitem{lattas2020avatarme}
Alexandros Lattas, Stylianos Moschoglou, Baris Gecer, Stylianos Ploumpis,
  Vasileios Triantafyllou, Abhijeet Ghosh, and Stefanos Zafeiriou.
\newblock Avatar{M}e: {R}ealistically renderable 3{D} facial reconstruction
  ··in-the-wild".
\newblock In {\em Proceedings of the IEEE/CVF Conference on Computer Vision and
  Pattern Recognition}, pages 760--769, 2020.

\bibitem{lattas2021avatarme++}
Alexandros Lattas, Stylianos Moschoglou, Stylianos Ploumpis, Baris Gecer,
  Abhijeet Ghosh, and Stefanos~P Zafeiriou.
\newblock Avatar{M}e++: {F}acial shape and {BRDF} inference with photorealistic
  rendering-aware {GAN}s.
\newblock {\em IEEE Transactions on Pattern Analysis \& Machine Intelligence},
  (01):1--1, 2021.

\bibitem{lee2020maskgan}
Cheng-Han Lee, Ziwei Liu, Lingyun Wu, and Ping Luo.
\newblock Mask{GAN}: {T}owards diverse and interactive facial image
  manipulation.
\newblock In {\em Proceedings of the IEEE/CVF conference on computer vision and
  pattern recognition}, pages 5549--5558, 2020.

\bibitem{lee2022rq-vae}
Doyup Lee, Chiheon Kim, Saehoon Kim, Minsu Cho, and Wook-Shin Han.
\newblock Autoregressive image generation using residual quantization.
\newblock In {\em Proceedings of the IEEE/CVF Conference on Computer Vision and
  Pattern Recognition}, pages 11523--11532, 2022.

\bibitem{li2019controllable}
Bowen Li, Xiaojuan Qi, Thomas Lukasiewicz, and Philip~HS Torr.
\newblock Controllable text-to-image generation.
\newblock In {\em Proceedings of the 33rd International Conference on Neural
  Information Processing Systems}, pages 2065--2075, 2019.

\bibitem{li2020manigan}
Bowen Li, Xiaojuan Qi, Thomas Lukasiewicz, and Philip~HS Torr.
\newblock Manigan: Text-guided image manipulation.
\newblock In {\em Proceedings of the IEEE/CVF Conference on Computer Vision and
  Pattern Recognition}, pages 7880--7889, 2020.

\bibitem{2020Lightweight}
Bowen Li, Xiaojuan Qi, Philip Torr, and Thomas Lukasiewicz.
\newblock Lightweight generative adversarial networks for text-guided image
  manipulation.
\newblock {\em Advances in Neural Information Processing Systems},
  33:22020--22031, 2020.

\bibitem{li2019visualbert}
Liunian~Harold Li, Mark Yatskar, Da Yin, Cho-Jui Hsieh, and Kai-Wei Chang.
\newblock Visualbert: A simple and performant baseline for vision and language.
\newblock Preprint arXiv:1908.03557, 2019.

\bibitem{flame}
Tianye Li, Timo Bolkart, Michael~J Black, Hao Li, and Javier Romero.
\newblock Learning a model of facial shape and expression from 4{D} scans.
\newblock {\em ACM Transactions on Graphics}, 36(6):194--1, 2017.

\bibitem{li2020unimo}
Wei Li, Can Gao, Guocheng Niu, Xinyan Xiao, Hao Liu, Jiachen Liu, Hua Wu, and
  Haifeng Wang.
\newblock {UNIMO}: {T}owards unified-modal understanding and generation via
  cross-modal contrastive learning.
\newblock {\em arXiv preprint arXiv:2012.15409}, 2020.

\bibitem{li2020oscar}
Xiujun Li, Xi Yin, Chunyuan Li, Pengchuan Zhang, Xiaowei Hu, Lei Zhang, Lijuan
  Wang, Houdong Hu, Li Dong, Furu Wei, et~al.
\newblock Oscar: {O}bject-semantics aligned pre-training for vision-language
  tasks.
\newblock In {\em European Conference on Computer Vision}, pages 121--137,
  2020.

\bibitem{lin2021m6}
Junyang Lin, Rui Men, An Yang, Chang Zhou, Ming Ding, Yichang Zhang, Peng Wang,
  Ang Wang, Le Jiang, Xianyan Jia, et~al.
\newblock M6: A chinese multimodal pretrainer.
\newblock Preprint arXiv:2103.00823, 2021.

\bibitem{mscoco}
Tsung-Yi Lin, Michael Maire, Serge Belongie, James Hays, Pietro Perona, Deva
  Ramanan, Piotr Doll{\'a}r, and C~Lawrence Zitnick.
\newblock Microsoft {coco}: {C}ommon objects in context.
\newblock In {\em European conference on computer vision}, pages 740--755,
  2014.

\bibitem{ssd}
Wei Liu, Dragomir Anguelov, Dumitru Erhan, Christian Szegedy, Scott Reed,
  Cheng-Yang Fu, and Alexander~C Berg.
\newblock {SSD}: {S}ingle shot multibox detector.
\newblock In {\em European conference on computer vision}, pages 21--37, 2016.

\bibitem{liu2022implicit-text-3D-generation}
Zhengzhe Liu, Yi Wang, Xiaojuan Qi, and Chi-Wing Fu.
\newblock Towards implicit text-guided 3d shape generation.
\newblock In {\em Proceedings of the IEEE/CVF Conference on Computer Vision and
  Pattern Recognition}, pages 17896--17906, 2022.

\bibitem{semantic_segmentation}
Jonathan Long, Evan Shelhamer, and Trevor Darrell.
\newblock Fully convolutional networks for semantic segmentation.
\newblock In {\em Proceedings of the IEEE/CVF conference on computer vision and
  pattern recognition}, pages 3431--3440, 2015.

\bibitem{lorensen1987marching}
William~E Lorensen and Harvey~E Cline.
\newblock Marching cubes: {A} high resolution 3{D} surface construction
  algorithm.
\newblock {\em ACM siggraph computer graphics}, 21(4):163--169, 1987.

\bibitem{rendering}
Nelson Max.
\newblock Optical models for direct volume rendering.
\newblock {\em IEEE Transactions on Visualization and Computer Graphics},
  1(2):99--108, 1995.

\bibitem{gan_r1}
Lars Mescheder, Andreas Geiger, and Sebastian Nowozin.
\newblock Which training methods for {GAN}s do actually converge?
\newblock In {\em International conference on machine learning}, pages
  3481--3490, 2018.

\bibitem{text2mesh}
Oscar Michel, Roi Bar-On, Richard Liu, Sagie Benaim, and Rana Hanocka.
\newblock Text2{M}esh: {T}ext-driven neural stylization for meshes.
\newblock In {\em Proceedings of the IEEE/CVF Conference on Computer Vision and
  Pattern Recognition}, pages 13492--13502, 2022.

\bibitem{mildenhall2020nerf}
Ben Mildenhall, Pratul~P Srinivasan, Matthew Tancik, Jonathan~T Barron, Ravi
  Ramamoorthi, and Ren Ng.
\newblock Nerf: Representing scenes as neural radiance fields for view
  synthesis.
\newblock In {\em European conference on computer vision}, pages 405--421,
  2020.

\bibitem{mirza2014conditional}
Mehdi Mirza and Simon Osindero.
\newblock Conditional generative adversarial nets.
\newblock {\em arXiv preprint arXiv:1411.1784}, 2014.

\bibitem{moschoglou20203dfacegan}
Stylianos Moschoglou, Stylianos Ploumpis, Mihalis~A Nicolaou, Athanasios
  Papaioannou, and Stefanos Zafeiriou.
\newblock {3DFaceGAN}: {A}dversarial nets for 3{D} face representation,
  generation, and translation.
\newblock {\em International Journal of Computer Vision}, 128(10):2534--2551,
  2020.

\bibitem{tagan}
Seonghyeon Nam, Yunji Kim, and Seon~Joo Kim.
\newblock Text-adaptive generative adversarial networks: {M}anipulating images
  with natural language.
\newblock In {\em Proceedings of the 32nd International Conference on Neural
  Information Processing Systems}, pages 42--51, 2018.

\bibitem{nasir2019text2facegan}
Osaid~Rehman Nasir, Shailesh~Kumar Jha, Manraj~Singh Grover, Yi Yu, Ajit Kumar,
  and Rajiv~Ratn Shah.
\newblock Text2{F}ace{GAN}: {F}ace generation from fine grained textual
  descriptions.
\newblock In {\em International Conference on Multimedia Big Data}, pages
  58--67, 2019.

\bibitem{ni2021m3p}
Minheng Ni, Haoyang Huang, Lin Su, Edward Cui, Taroon Bharti, Lijuan Wang,
  Dongdong Zhang, and Nan Duan.
\newblock {M3P}: {L}earning universal representations via multitask
  multilingual multimodal pre-training.
\newblock In {\em Proceedings of the IEEE/CVF Conference on Computer Vision and
  Pattern Recognition}, pages 3977--3986, 2021.

\bibitem{nichol2021glide}
Alex Nichol, Prafulla Dhariwal, Aditya Ramesh, Pranav Shyam, Pamela Mishkin,
  Bob McGrew, Ilya Sutskever, and Mark Chen.
\newblock Glide: Towards photorealistic image generation and editing with
  text-guided diffusion models.
\newblock {\em arXiv preprint arXiv:2112.10741}, 2021.

\bibitem{oxford102}
Maria-Elena Nilsback and Andrew Zisserman.
\newblock Automated flower classification over a large number of classes.
\newblock In {\em Indian Conference on Computer Vision, Graphics \& Image
  Processing}, pages 722--729, 2008.

\bibitem{park2019SPADE}
Taesung Park, Ming-Yu Liu, Ting-Chun Wang, and Jun-Yan Zhu.
\newblock Semantic image synthesis with spatially-adaptive normalization.
\newblock In {\em Proceedings of the IEEE/CVF Conference on Computer Vision and
  Pattern Recognition}, pages 2337--2346, 2019.

\bibitem{patashnik2021styleclip}
Or Patashnik, Zongze Wu, Eli Shechtman, Daniel Cohen-Or, and Dani Lischinski.
\newblock Style{CLIP}: {T}ext-driven manipulation of stylegan imagery.
\newblock In {\em Proceedings of the IEEE/CVF International Conference on
  Computer Vision}, pages 2085--2094, 2021.

\bibitem{pixelface}
Jun Peng, Xiaoxiong Du, Yiyi Zhou, Jing He, Yunhang Shen, Xiaoshuai Sun, and
  Rongrong Ji.
\newblock Learning dynamic prior knowledge for text-to-face pixel synthesis.
\newblock In {\em Proceedings of the ACM International Conference on
  Multimedia}, pages 5132--5141, 2022.

\bibitem{openfacegan}
Jun Peng, Han Pan, Yiyi Zhou, Jing He, Xiaoshuai Sun, Yan Wang, Yongjian Wu,
  and Rongrong Ji.
\newblock Towards open-ended text-to-face generation, combination and
  manipulation.
\newblock In {\em Proceedings of the ACM International Conference on
  Multimedia}, pages 5045--5054, 2022.

\bibitem{frgc}
P~Jonathon Phillips, Patrick~J Flynn, Todd Scruggs, Kevin~W Bowyer, Jin Chang,
  Kevin Hoffman, Joe Marques, Jaesik Min, and William Worek.
\newblock Overview of the face recognition grand challenge.
\newblock In {\em Proceedings of the IEEE/CVF conference on computer vision and
  pattern recognition}, volume~1, pages 947--954, 2005.

\bibitem{poole2022dreamfusion}
Ben Poole, Ajay Jain, Jonathan~T Barron, and Ben Mildenhall.
\newblock Dream{F}usion: {T}ext-to-3{D} using 2{D} diffusion.
\newblock {\em arXiv preprint arXiv:2209.14988}, 2022.

\bibitem{entity_segmentation}
Lu Qi, Jason Kuen, Yi Wang, Jiuxiang Gu, Hengshuang Zhao, Zhe Lin, Philip Torr,
  and Jiaya Jia.
\newblock Open-world entity segmentation.
\newblock {\em arXiv preprint arXiv:2107.14228}, 2021.

\bibitem{radford2021clip}
Alec Radford, Jong~Wook Kim, Chris Hallacy, Aditya Ramesh, Gabriel Goh,
  Sandhini Agarwal, Girish Sastry, Amanda Askell, Pamela Mishkin, Jack Clark,
  et~al.
\newblock Learning transferable visual models from natural language
  supervision.
\newblock In {\em International Conference on Machine Learning}, pages
  8748--8763, 2021.

\bibitem{ramesh2022dalle2}
Aditya Ramesh, Prafulla Dhariwal, Alex Nichol, Casey Chu, and Mark Chen.
\newblock Hierarchical text-conditional image generation with clip latents.
\newblock {\em arXiv preprint arXiv:2204.06125}, 2022.

\bibitem{dalle}
Aditya Ramesh, Mikhail Pavlov, Gabriel Goh, Scott Gray, Chelsea Voss, Alec
  Radford, Mark Chen, and Ilya Sutskever.
\newblock Zero-shot text-to-image generation.
\newblock In {\em International Conference on Machine Learning}, pages
  8821--8831, 2021.

\bibitem{vqvae2}
Ali Razavi, Aaron van~den Oord, and Oriol Vinyals.
\newblock Generating diverse high-fidelity images with {VQ-VAE}-2.
\newblock In {\em Advances in Neural Information Processing Systems}, pages
  14866--14876, 2019.

\bibitem{yolo}
Joseph Redmon, Santosh Divvala, Ross Girshick, and Ali Farhadi.
\newblock You only look once: {U}nified, real-time object detection.
\newblock In {\em Proceedings of the IEEE/CVF conference on computer vision and
  pattern recognition}, pages 779--788, 2016.

\bibitem{GAN-INT-CLS}
Scott Reed, Zeynep Akata, Xinchen Yan, Lajanugen Logeswaran, Bernt Schiele, and
  Honglak Lee.
\newblock Generative adversarial text to image synthesis.
\newblock In {\em International Conference on Machine Learning}, pages
  1060--1069, 2016.

\bibitem{reizenstein2021common}
Jeremy Reizenstein, Roman Shapovalov, Philipp Henzler, Luca Sbordone, Patrick
  Labatut, and David Novotny.
\newblock Common objects in 3{D}: Large-scale learning and evaluation of
  real-life 3{D} category reconstruction.
\newblock In {\em Proceedings of the IEEE/CVF International Conference on
  Computer Vision}, pages 10901--10911, 2021.

\bibitem{fasterrcnn}
Shaoqing Ren, Kaiming He, Ross Girshick, and Jian Sun.
\newblock Faster {R-CNN}: {T}owards real-time object detection with region
  proposal networks.
\newblock {\em Advances in Neural Information Processing Systems}, 28:91--99,
  2015.

\bibitem{roich2022pivotal}
Daniel Roich, Ron Mokady, Amit~H Bermano, and Daniel Cohen-Or.
\newblock Pivotal tuning for latent-based editing of real images.
\newblock {\em ACM Transactions on Graphics}, 42(1):1--13, 2022.

\bibitem{latent-diffusion}
Robin Rombach, Andreas Blattmann, Dominik Lorenz, Patrick Esser, and Bj{\"o}rn
  Ommer.
\newblock High-resolution image synthesis with latent diffusion models.
\newblock In {\em Proceedings of the IEEE/CVF Conference on Computer Vision and
  Pattern Recognition}, pages 10684--10695, 2022.

\bibitem{imagen}
Chitwan Saharia, William Chan, Saurabh Saxena, Lala Li, Jay Whang, Emily
  Denton, Seyed Kamyar~Seyed Ghasemipour, Burcu~Karagol Ayan, S~Sara Mahdavi,
  Rapha~Gontijo Lopes, et~al.
\newblock Photorealistic text-to-image diffusion models with deep language
  understanding.
\newblock {\em arXiv preprint arXiv:2205.11487}, 2022.

\bibitem{salimans2016IS}
Tim Salimans, Ian Goodfellow, Wojciech Zaremba, Vicki Cheung, Alec Radford, and
  Xi Chen.
\newblock Improved techniques for training gans.
\newblock {\em Advances in Neural Information Processing Systems},
  29:2234--2242, 2016.

\bibitem{sanghi2021clip-forge}
Aditya Sanghi, Hang Chu, Joseph~G Lambourne, Ye Wang, Chin-Yi Cheng, Marco
  Fumero, and Kamal~Rahimi Malekshan.
\newblock {CLIP}-{F}orge: Towards zero-shot text-to-shape generation.
\newblock In {\em Proceedings of the IEEE/CVF Conference on Computer Vision and
  Pattern Recognition}, pages 18603--18613, 2022.

\bibitem{schonberger2016structure}
Johannes~L Schonberger and Jan-Michael Frahm.
\newblock Structure-from-motion revisited.
\newblock In {\em Proceedings of the IEEE/CVF conference on computer vision and
  pattern recognition}, pages 4104--4113, 2016.

\bibitem{facenet}
Florian Schroff, Dmitry Kalenichenko, and James Philbin.
\newblock Face{N}et: {A} unified embedding for face recognition and clustering.
\newblock In {\em Proceedings of the IEEE/CVF conference on computer vision and
  pattern recognition}, pages 815--823, 2015.

\bibitem{schwarz2020graf}
Katja Schwarz, Yiyi Liao, Michael Niemeyer, and Andreas Geiger.
\newblock {GRAF}: {Ge}nerative radiance fields for 3{D}-aware image synthesis.
\newblock {\em Advances in Neural Information Processing Systems},
  33:20154--20166, 2020.

\bibitem{sennrich2015neural}
Rico Sennrich, Barry Haddow, and Alexandra Birch.
\newblock Neural machine translation of rare words with subword units.
\newblock {\em arXiv preprint arXiv:1508.07909}, 2015.

\bibitem{sennrich2015bpe}
Rico Sennrich, Barry Haddow, and Alexandra Birch.
\newblock Neural machine translation of rare words with subword units.
\newblock {\em arXiv preprint arXiv:1508.07909}, 2015.

\bibitem{shazeer2020glu}
Noam Shazeer.
\newblock Glu variants improve transformer.
\newblock {\em arXiv preprint arXiv:2002.05202}, 2020.

\bibitem{siddiqui2022texturify}
Yawar Siddiqui, Justus Thies, Fangchang Ma, Qi Shan, Matthias Nie{\ss}ner, and
  Angela Dai.
\newblock Texturify: {G}enerating textures on 3{D} shape surfaces.
\newblock {\em arXiv preprint arXiv:2204.02411}, 2022.

\bibitem{stap_Conditional}
David Stap, Maurits Bleeker, Sarah Ibrahimi, and Maartje ter Hoeve.
\newblock Conditional image generation and manipulation for user-specified
  content.
\newblock {\em arXiv preprint arXiv:2005.04909}, 2020.

\bibitem{su2019vlBERT}
Weijie Su, Xizhou Zhu, Yue Cao, Bin Li, Lewei Lu, Furu Wei, and Jifeng Dai.
\newblock {VL-BERT}: {P}re-training of generic visual-linguistic
  representations.
\newblock In {\em International Conference on Learning Representations}, 2019.

\bibitem{anyface}
Jianxin Sun, Qiyao Deng, Qi Li, Muyi Sun, Min Ren, and Zhenan Sun.
\newblock Any{F}ace: {F}ree-style text-to-face synthesis and manipulation.
\newblock In {\em Proceedings of the IEEE/CVF Conference on Computer Vision and
  Pattern Recognition}, pages 18687--18696, 2022.

\bibitem{sea-t2f}
Jianxin Sun, Qi Li, Weining Wang, Jian Zhao, and Zhenan Sun.
\newblock Multi-caption text-to-face synthesis: {D}ataset and algorithm.
\newblock In {\em Proceedings of the ACM International Conference on
  Multimedia}, pages 2290--2298, 2021.

\bibitem{sun2022fenerf}
Jingxiang Sun, Xuan Wang, Yong Zhang, Xiaoyu Li, Qi Zhang, Yebin Liu, and Jue
  Wang.
\newblock {FE}{N}e{RF}: {F}ace editing in neural radiance fields.
\newblock In {\em Proceedings of the IEEE/CVF Conference on Computer Vision and
  Pattern Recognition}, pages 7672--7682, 2022.

\bibitem{tan2019lxmert}
Hao Tan and Mohit Bansal.
\newblock Lxmert: {L}earning cross-modality encoder representations from
  transformers.
\newblock {\em arXiv preprint arXiv:1908.07490}, 2019.

\bibitem{tao2020dfgan}
Ming Tao, Hao Tang, Songsong Wu, Nicu Sebe, Xiao-Yuan Jing, Fei Wu, and Bingkun
  Bao.
\newblock {DF-GAN}: {D}eep fusion generative adversarial networks for
  text-to-image synthesis.
\newblock {\em arXiv preprint arXiv:2008.05865}, 2020.

\bibitem{2017Neural}
Aaron Van Den~Oord, Oriol Vinyals, et~al.
\newblock Neural discrete representation learning.
\newblock {\em Advances in Neural Information Processing Systems}, 30, 2017.

\bibitem{vaswani2017attention}
Ashish Vaswani, Noam Shazeer, Niki Parmar, Jakob Uszkoreit, Llion Jones,
  Aidan~N Gomez, {\L}ukasz Kaiser, and Illia Polosukhin.
\newblock Attention is all you need.
\newblock In {\em Advances in Neural Information Processing Systems}, pages
  5998--6008, 2017.

\bibitem{cub200}
C. Wah, S. Branson, P. Welinder, P. Perona, and S. Belongie.
\newblock {The Caltech-UCSD Birds-200-2011 Dataset}.
\newblock Technical Report CNS-TR-2011-001, California Institute of Technology,
  2011.

\bibitem{wang2021clip}
Can Wang, Menglei Chai, Mingming He, Dongdong Chen, and Jing Liao.
\newblock {CLIP-N}e{RF}: {T}ext-and-image driven manipulation of neural
  radiance fields.
\newblock In {\em Proceedings of the IEEE/CVF Conference on Computer Vision and
  Pattern Recognition}, pages 3835--3844, 2022.

\bibitem{wang2018learning}
Hai Wang, Jason~D Williams, and SingBing Kang.
\newblock Learning to globally edit images with textual description.
\newblock {\em arXiv preprint arXiv:1810.05786}, 2018.

\bibitem{faces-a-la-carte}
Tianren Wang, Teng Zhang, and Brian Lovell.
\newblock Faces a la carte: {T}ext-to-face generation via attribute
  disentanglement.
\newblock In {\em Proceedings of the IEEE/CVF winter conference on applications
  of computer vision}, pages 3380--3388, 2021.

\bibitem{wang2018pix2pixhd}
Ting-Chun Wang, Ming-Yu Liu, Jun-Yan Zhu, Andrew Tao, Jan Kautz, and Bryan
  Catanzaro.
\newblock High-resolution image synthesis and semantic manipulation with
  conditional gans.
\newblock In {\em Proceedings of the IEEE/CVF conference on computer vision and
  pattern recognition}, pages 8798--8807, 2018.

\bibitem{wang2004image}
Zhou Wang, Alan~C Bovik, Hamid~R Sheikh, and Eero~P Simoncelli.
\newblock Image quality assessment: {F}rom error visibility to structural
  similarity.
\newblock {\em IEEE transactions on image processing}, 13(4):600--612, 2004.

\bibitem{wang2021simvlm}
Zirui Wang, Jiahui Yu, Adams~Wei Yu, Zihang Dai, Yulia Tsvetkov, and Yuan Cao.
\newblock Simvlm: Simple visual language model pretraining with weak
  supervision.
\newblock Preprint arXiv:2108.10904, 2021.

\bibitem{xia2021tedigan}
Weihao Xia, Yujiu Yang, Jing-Hao Xue, and Baoyuan Wu.
\newblock Tedi{GAN}: {T}ext-guided diverse face image generation and
  manipulation.
\newblock In {\em Proceedings of the IEEE/CVF Conference on Computer Vision and
  Pattern Recognition}, pages 2256--2265, 2021.

\bibitem{tedigan-b}
Weihao Xia, Yujiu Yang, Jing-Hao Xue, and Baoyuan Wu.
\newblock Towards open-world text-guided face image generation and
  manipulation.
\newblock {\em arXiv preprint arXiv:2104.08910}, 2021.

\bibitem{xu2018attngan}
Tao Xu, Pengchuan Zhang, Qiuyuan Huang, Han Zhang, Zhe Gan, Xiaolei Huang, and
  Xiaodong He.
\newblock Attn{GAN}: {F}ine-grained text to image generation with attentional
  generative adversarial networks.
\newblock In {\em Proceedings of the IEEE/CVF conference on computer vision and
  pattern recognition}, pages 1316--1324, 2018.

\bibitem{yao2021filip}
Lewei Yao, Runhui Huang, Lu Hou, Guansong Lu, Minzhe Niu, Hang Xu, Xiaodan
  Liang, Zhenguo Li, Xin Jiang, and Chunjing Xu.
\newblock Filip: Fine-grained interactive language-image pre-training, 2021.

\bibitem{ye2021xmcgan}
Hui Ye, Xiulong Yang, Martin Takac, Rajshekhar Sunderraman, and Shihao Ji.
\newblock Improving text-to-image synthesis using contrastive learning.
\newblock {\em arXiv preprint arXiv:2107.02423}, 2021.

\bibitem{yi2017dualgan}
Zili Yi, Hao Zhang, Ping Tan, and Minglun Gong.
\newblock Dualgan: Unsupervised dual learning for image-to-image translation.
\newblock In {\em Proceedings of the IEEE international conference on computer
  vision}, pages 2849--2857, 2017.

\bibitem{yu2021pixelnerf}
Alex Yu, Vickie Ye, Matthew Tancik, and Angjoo Kanazawa.
\newblock pixelnerf: Neural radiance fields from one or few images.
\newblock In {\em Proceedings of the IEEE/CVF Conference on Computer Vision and
  Pattern Recognition}, pages 4578--4587, 2021.

\bibitem{yu2018bisenet}
Changqian Yu, Jingbo Wang, Chao Peng, Changxin Gao, Gang Yu, and Nong Sang.
\newblock Bi{S}e{N}et: {B}ilateral segmentation network for real-time semantic
  segmentation.
\newblock In {\em Proceedings of the European conference on computer vision},
  pages 325--341, 2018.

\bibitem{parti}
Jiahui Yu, Yuanzhong Xu, Jing~Yu Koh, Thang Luong, Gunjan Baid, Zirui Wang,
  Vijay Vasudevan, Alexander Ku, Yinfei Yang, Burcu~Karagol Ayan, et~al.
\newblock Scaling autoregressive models for content-rich text-to-image
  generation.
\newblock {\em arXiv preprint arXiv:2206.10789}, 2022.

\bibitem{zhang2017stackgan}
Han Zhang, Tao Xu, Hongsheng Li, Shaoting Zhang, Xiaogang Wang, Xiaolei Huang,
  and Dimitris~N Metaxas.
\newblock Stack{GAN}: {T}ext to photo-realistic image synthesis with stacked
  generative adversarial networks.
\newblock In {\em Proceedings of the IEEE international conference on computer
  vision}, pages 5907--5915, 2017.

\bibitem{zhang2018stackgan++}
Han Zhang, Tao Xu, Hongsheng Li, Shaoting Zhang, Xiaogang Wang, Xiaolei Huang,
  and Dimitris~N Metaxas.
\newblock Stack{GAN}++: {R}ealistic image synthesis with stacked generative
  adversarial networks.
\newblock {\em IEEE transactions on pattern analysis and machine intelligence},
  41(8):1947--1962, 2018.

\bibitem{zhang2021vinvl}
Pengchuan Zhang, Xiujun Li, Xiaowei Hu, Jianwei Yang, Lei Zhang, Lijuan Wang,
  Yejin Choi, and Jianfeng Gao.
\newblock Vin{VL}: {R}evisiting visual representations in vision-language
  models.
\newblock In {\em Proceedings of the IEEE/CVF Conference on Computer Vision and
  Pattern Recognition}, pages 5579--5588, 2021.

\bibitem{zhang2018unreasonable}
Richard Zhang, Phillip Isola, Alexei~A Efros, Eli Shechtman, and Oliver Wang.
\newblock The unreasonable effectiveness of deep features as a perceptual
  metric.
\newblock In {\em Proceedings of the IEEE/CVF conference on computer vision and
  pattern recognition}, pages 586--595, 2018.

\bibitem{lpips}
Richard Zhang, Phillip Isola, Alexei~A Efros, Eli Shechtman, and Oliver Wang.
\newblock The unreasonable effectiveness of deep features as a perceptual
  metric.
\newblock In {\em Proceedings of the IEEE/CVF conference on computer vision and
  pattern recognition}, pages 586--595, 2018.

\bibitem{zhang2021text}
Tianhao Zhang, Hung-Yu Tseng, Lu Jiang, Weilong Yang, Honglak Lee, and Irfan
  Essa.
\newblock Text as neural operator: Image manipulation by text instruction.
\newblock In {\em Proceedings of the ACM International Conference on
  Multimedia}, pages 1893--1902, 2021.

\bibitem{zhang2021m6-ufc}
Zhu Zhang, Jianxin Ma, Chang Zhou, Rui Men, Zhikang Li, Ming Ding, Jie Tang,
  Jingren Zhou, and Hongxia Yang.
\newblock M6-{UFC}: {U}nifying multi-modal controls for conditional image
  synthesis.
\newblock {\em arXiv preprint arXiv:2105.14211}, 2021.

\bibitem{zhou20213d}
Linqi Zhou, Yilun Du, and Jiajun Wu.
\newblock 3{D} shape generation and completion through point-voxel diffusion.
\newblock In {\em Proceedings of the IEEE/CVF International Conference on
  Computer Vision}, pages 5826--5835, 2021.

\bibitem{zhou2021uc2}
Mingyang Zhou, Luowei Zhou, Shuohang Wang, Yu Cheng, Linjie Li, Zhou Yu, and
  Jingjing Liu.
\newblock {UC2}: {U}niversal cross-lingual cross-modal vision-and-language
  pre-training.
\newblock In {\em Proceedings of the IEEE/CVF Conference on Computer Vision and
  Pattern Recognition}, pages 4155--4165, 2021.

\bibitem{FFHQ_text}
Yutong Zhou.
\newblock Generative adversarial network for text-to-face synthesis and
  manipulation.
\newblock In {\em Proceedings of the ACM International Conference on
  Multimedia}, pages 2940--2944, 2021.

\bibitem{zhu2017cyclegan}
Jun-Yan Zhu, Taesung Park, Phillip Isola, and Alexei~A Efros.
\newblock Unpaired image-to-image translation using cycle-consistent
  adversarial networks.
\newblock In {\em Proceedings of the IEEE international conference on computer
  vision}, pages 2223--2232, 2017.

\bibitem{zhu2019dmgan}
Minfeng Zhu, Pingbo Pan, Wei Chen, and Yi Yang.
\newblock {DM-GAN}: {D}ynamic memory generative adversarial networks for
  text-to-image synthesis.
\newblock In {\em Proceedings of the IEEE/CVF Conference on Computer Vision and
  Pattern Recognition}, pages 5802--5810, 2019.

\bibitem{zhuge2021kaleido}
Mingchen Zhuge, Dehong Gao, Deng-Ping Fan, Linbo Jin, Ben Chen, Haoming Zhou,
  Minghui Qiu, and Ling Shao.
\newblock Kaleido-{BERT}: {V}ision-language pre-training on fashion domain.
\newblock In {\em Proceedings of the IEEE/CVF Conference on Computer Vision and
  Pattern Recognition}, pages 12647--12657, 2021.

\end{thebibliography}


\begin{thebibliography}{100}\itemsep=-1pt

\bibitem{aneja2023clipface}
Shivangi Aneja, Justus Thies, Angela Dai, and Matthias Nie{\ss}ner.
\newblock Clip{F}ace: {T}ext-guided editing of textured 3{D} morphable models.
\newblock In {\em SIGGRAPH '23 Conference Proceedings}, 2023.


\bibitem{athar2022rignerf}
ShahRukh Athar, Zexiang Xu, Kalyan Sunkavalli, Eli Shechtman, and Zhixin Shu.
\newblock Rig{N}e{RF}: {F}ully controllable neural 3{D} portraits.
\newblock In {\em Proceedings of the IEEE/CVF Conference on Computer Vision and
  Pattern Recognition}, pages 20364--20373, 2022.


\bibitem{canfes2022text}
Zehranaz Canfes, M~Furkan Atasoy, Alara Dirik, and Pinar Yanardag.
\newblock Text and image guided 3{D} avatar generation and manipulation.
\newblock {\em arXiv preprint arXiv:2202.06079}, 2022.


\bibitem{eg3d}
Eric~R Chan, Connor~Z Lin, Matthew~A Chan, Koki Nagano, Boxiao Pan, Shalini
  De~Mello, Orazio Gallo, Leonidas~J Guibas, Jonathan Tremblay, Sameh Khamis,
  et~al.
\newblock Efficient geometry-aware 3{D} generative adversarial networks.
\newblock In {\em Proceedings of the IEEE/CVF Conference on Computer Vision and
  Pattern Recognition}, pages 16123--16133, 2022.

\bibitem{pigan}
Eric~R Chan, Marco Monteiro, Petr Kellnhofer, Jiajun Wu, and Gordon Wetzstein.
\newblock Pi-{GAN}: {P}eriodic implicit generative adversarial networks for
  3{D}-aware image synthesis.
\newblock In {\em Proceedings of the IEEE/CVF Conference on Computer Vision and
  Pattern Recognition}, pages 5799--5809, 2021.


\bibitem{chen2018text2shape}
Kevin Chen, Christopher~B Choy, Manolis Savva, Angel~X Chang, Thomas
  Funkhouser, and Silvio Savarese.
\newblock Text2shape: Generating shapes from natural language by learning joint
  embeddings.
\newblock In {\em Asian conference on computer vision}, pages 100--116, 2018.



\bibitem{ding2021cogview}
Ming Ding, Zhuoyi Yang, Wenyi Hong, Wendi Zheng, Chang Zhou, Da Yin, Junyang
  Lin, Xu Zou, Zhou Shao, Hongxia Yang, et~al.
\newblock Cog{V}iew: {M}astering text-to-image generation via transformers.
\newblock {\em Advances in Neural Information Processing Systems},
  34:19822--19835, 2021.

\bibitem{ding2022cogview2}
Ming Ding, Wendi Zheng, Wenyi Hong, and Jie Tang.
\newblock Cog{V}iew2: Faster and better text-to-image generation via
  hierarchical transformers.
\newblock {\em arXiv preprint arXiv:2204.14217}, 2022.

\bibitem{sisgan}
Hao Dong, Simiao Yu, Chao Wu, and Yike Guo.
\newblock Semantic image synthesis via adversarial learning.
\newblock In {\em Proceedings of the IEEE International Conference on Computer
  Vision}, pages 5706--5714, 2017.


\bibitem{Ekman1978FacialAC}
Paul Ekman and Wallace~V. Friesen.
\newblock Facial action coding system: {A} technique for the measurement of
  facial movement.
\newblock 1978.


\bibitem{esser2021imagebart}
Patrick Esser, Robin Rombach, Andreas Blattmann, and Bjorn Ommer.
\newblock Image{BART}: {B}idirectional context with multinomial diffusion for
  autoregressive image synthesis.
\newblock {\em Advances in Neural Information Processing Systems},
  34:3518--3532, 2021.


\bibitem{fu2022shapecrafter}
Rao Fu, Xiao Zhan, Yiwen Chen, Daniel Ritchie, and Srinath Sridhar.
\newblock Shape{C}rafter: {A} recursive text-conditioned 3{D} shape generation
  model.
\newblock {\em arXiv preprint arXiv:2207.09446}, 2022.


\bibitem{goodfellow2014gan}
Ian Goodfellow, Jean Pouget-Abadie, Mehdi Mirza, Bing Xu, David Warde-Farley,
  Sherjil Ozair, Aaron Courville, and Yoshua Bengio.
\newblock Generative adversarial nets.
\newblock {\em Advances in Neural Information Processing Systems}, 27, 2014.

\bibitem{heusel2017FID}
Martin Heusel, Hubert Ramsauer, Thomas Unterthiner, Bernhard Nessler, and Sepp
  Hochreiter.
\newblock {GAN}s trained by a two time-scale update rule converge to a local
  nash equilibrium.
\newblock {\em Advances in Neural Information Processing Systems}, 30, 2017.

\bibitem{ho2020ddpm}
Jonathan Ho, Ajay Jain, and Pieter Abbeel.
\newblock Denoising diffusion probabilistic models.
\newblock {\em Advances in Neural Information Processing Systems},
  33:6840--6851, 2020.


\bibitem{hong2022avatarclip}
Fangzhou Hong, Mingyuan Zhang, Liang Pan, Zhongang Cai, Lei Yang, and Ziwei
  Liu.
\newblock Avatar{CLIP}: {Z}ero-shot text-driven generation and animation of
  3{D} avatars.
\newblock {\em ACM Transactions on Graphics}, 41(4):1--19, 2022.

\bibitem{hu2021text}
Li Hu, Jinwei Qi, Bang Zhang, Pan Pan, and Yinghui Xu.
\newblock Text-driven 3{D} avatar animation with emotional and expressive
  behaviors.
\newblock In {\em Proceedings of the ACM International Conference on
  Multimedia}, pages 2816--2818, 2021.


\bibitem{jain2021dreamfields}
Ajay Jain, Ben Mildenhall, Jonathan~T Barron, Pieter Abbeel, and Ben Poole.
\newblock Zero-shot text-guided object generation with dream fields.
\newblock In {\em Proceedings of the IEEE/CVF Conference on Computer Vision and
  Pattern Recognition}, pages 867--876, 2022.

\bibitem{jetchev2021clipmatrix}
Nikolay Jetchev.
\newblock Clip{M}atrix: {T}ext-controlled creation of 3{D} textured meshes.
\newblock {\em arXiv preprint arXiv:2109.12922}, 2021.


\bibitem{stylegan}
Tero Karras, Samuli Laine, and Timo Aila.
\newblock A style-based generator architecture for generative adversarial
  networks.
\newblock In {\em Proceedings of the IEEE/CVF Conference on Computer Vision and
  Pattern Recognition}, pages 4401--4410, 2019.

\bibitem{khalid2022clip-mesh}
Nasir Khalid, Tianhao Xie, Eugene Belilovsky, and Tiberiu Popa.
\newblock {CLIP}-{M}esh: {G}enerating textured meshes from text using
  pretrained image-text models.
\newblock {\em ACM Transactions on Graphics}, 2022.


\bibitem{lattas2020avatarme}
Alexandros Lattas, Stylianos Moschoglou, Baris Gecer, Stylianos Ploumpis,
  Vasileios Triantafyllou, Abhijeet Ghosh, and Stefanos Zafeiriou.
\newblock Avatar{M}e: {R}ealistically renderable 3{D} facial reconstruction in-the-wild".
\newblock In {\em Proceedings of the IEEE/CVF Conference on Computer Vision and
  Pattern Recognition}, pages 760--769, 2020.

\bibitem{lattas2021avatarme++}
Alexandros Lattas, Stylianos Moschoglou, Stylianos Ploumpis, Baris Gecer,
  Abhijeet Ghosh, and Stefanos~P Zafeiriou.
\newblock Avatar{M}e++: {F}acial shape and {BRDF} inference with photorealistic
  rendering-aware {GAN}s.
\newblock {\em IEEE Transactions on Pattern Analysis \& Machine Intelligence},
  (01):1--1, 2021.

\bibitem{lee2020maskgan}
Cheng-Han Lee, Ziwei Liu, Lingyun Wu, and Ping Luo.
\newblock Mask{GAN}: {T}owards diverse and interactive facial image
  manipulation.
\newblock In {\em Proceedings of the IEEE/CVF conference on computer vision and
  pattern recognition}, pages 5549--5558, 2020.

\bibitem{lee2022rq-vae}
Doyup Lee, Chiheon Kim, Saehoon Kim, Minsu Cho, and Wook-Shin Han.
\newblock Autoregressive image generation using residual quantization.
\newblock In {\em Proceedings of the IEEE/CVF Conference on Computer Vision and
  Pattern Recognition}, pages 11523--11532, 2022.

\bibitem{li2019controllable}
Bowen Li, Xiaojuan Qi, Thomas Lukasiewicz, and Philip~HS Torr.
\newblock Controllable text-to-image generation.
\newblock In {\em Proceedings of the 33rd International Conference on Neural
  Information Processing Systems}, pages 2065--2075, 2019.



\bibitem{flame}
Tianye Li, Timo Bolkart, Michael~J Black, Hao Li, and Javier Romero.
\newblock Learning a model of facial shape and expression from 4{D} scans.
\newblock {\em ACM Transactions on Graphics}, 36(6):194--1, 2017.




\bibitem{liu2022implicit-text-3D-generation}
Zhengzhe Liu, Yi Wang, Xiaojuan Qi, and Chi-Wing Fu.
\newblock Towards implicit text-guided 3{D} shape generation.
\newblock In {\em Proceedings of the IEEE/CVF Conference on Computer Vision and
  Pattern Recognition}, pages 17896--17906, 2022.


\bibitem{rendering}
Nelson Max.
\newblock Optical models for direct volume rendering.
\newblock {\em IEEE Transactions on Visualization and Computer Graphics},
  1(2):99--108, 1995.


\bibitem{text2mesh}
Oscar Michel, Roi Bar-On, Richard Liu, Sagie Benaim, and Rana Hanocka.
\newblock Text2{M}esh: {T}ext-driven neural stylization for meshes.
\newblock In {\em Proceedings of the IEEE/CVF Conference on Computer Vision and
  Pattern Recognition}, pages 13492--13502, 2022.


\bibitem{moschoglou20203dfacegan}
Stylianos Moschoglou, Stylianos Ploumpis, Mihalis~A Nicolaou, Athanasios
  Papaioannou, and Stefanos Zafeiriou.
\newblock {3DFaceGAN}: {A}dversarial nets for 3{D} face representation,
  generation, and translation.
\newblock {\em International Journal of Computer Vision}, 128(10):2534--2551,
  2020.


\bibitem{nasir2019text2facegan}
Osaid~Rehman Nasir, Shailesh~Kumar Jha, Manraj~Singh Grover, Yi Yu, Ajit Kumar,
  and Rajiv~Ratn Shah.
\newblock Text2{F}ace{GAN}: {F}ace generation from fine grained textual
  descriptions.
\newblock In {\em International Conference on Multimedia Big Data}, pages
  58--67, 2019.


\bibitem{nichol2021glide}
Alex Nichol, Prafulla Dhariwal, Aditya Ramesh, Pranav Shyam, Pamela Mishkin,
  Bob McGrew, Ilya Sutskever, and Mark Chen.
\newblock Glide: Towards photorealistic image generation and editing with
  text-guided diffusion models.
\newblock {\em arXiv preprint arXiv:2112.10741}, 2021.

\bibitem{pixelface}
Jun Peng, Xiaoxiong Du, Yiyi Zhou, Jing He, Yunhang Shen, Xiaoshuai Sun, and
  Rongrong Ji.
\newblock Learning dynamic prior knowledge for text-to-face pixel synthesis.
\newblock In {\em Proceedings of the ACM International Conference on
  Multimedia}, pages 5132--5141, 2022.

\bibitem{openfacegan}
Jun Peng, Han Pan, Yiyi Zhou, Jing He, Xiaoshuai Sun, Yan Wang, Yongjian Wu,
  and Rongrong Ji.
\newblock Towards open-ended text-to-face generation, combination and
  manipulation.
\newblock In {\em Proceedings of the ACM International Conference on
  Multimedia}, pages 5045--5054, 2022.

\bibitem{poole2022dreamfusion}
Ben Poole, Ajay Jain, Jonathan~T Barron, and Ben Mildenhall.
\newblock Dream{F}usion: {T}ext-to-3{D} using 2{D} diffusion.
\newblock {\em arXiv preprint arXiv:2209.14988}, 2022.


\bibitem{ramesh2022dalle2}
Aditya Ramesh, Prafulla Dhariwal, Alex Nichol, Casey Chu, and Mark Chen.
\newblock Hierarchical text-conditional image generation with clip latents.
\newblock {\em arXiv preprint arXiv:2204.06125}, 2022.

\bibitem{dalle}
Aditya Ramesh, Mikhail Pavlov, Gabriel Goh, Scott Gray, Chelsea Voss, Alec
  Radford, Mark Chen, and Ilya Sutskever.
\newblock Zero-shot text-to-image generation.
\newblock In {\em International Conference on Machine Learning}, pages
  8821--8831, 2021.


\bibitem{GAN-INT-CLS}
Scott Reed, Zeynep Akata, Xinchen Yan, Lajanugen Logeswaran, Bernt Schiele, and
  Honglak Lee.
\newblock Generative adversarial text to image synthesis.
\newblock In {\em International Conference on Machine Learning}, pages
  1060--1069, 2016.

\bibitem{roich2022pivotal}
Daniel Roich, Ron Mokady, Amit~H Bermano, and Daniel Cohen-Or.
\newblock Pivotal tuning for latent-based editing of real images.
\newblock {\em ACM Transactions on Graphics}, 42(1):1--13, 2022.

\bibitem{latent-diffusion}
Robin Rombach, Andreas Blattmann, Dominik Lorenz, Patrick Esser, and Bj{\"o}rn
  Ommer.
\newblock High-resolution image synthesis with latent diffusion models.
\newblock In {\em Proceedings of the IEEE/CVF Conference on Computer Vision and
  Pattern Recognition}, pages 10684--10695, 2022.

\bibitem{imagen}
Chitwan Saharia, William Chan, Saurabh Saxena, Lala Li, Jay Whang, Emily
  Denton, Seyed Kamyar~Seyed Ghasemipour, Burcu~Karagol Ayan, S~Sara Mahdavi,
  Rapha~Gontijo Lopes, et~al.
\newblock Photorealistic text-to-image diffusion models with deep language
  understanding.
\newblock {\em arXiv preprint arXiv:2205.11487}, 2022.


\bibitem{sanghi2021clip-forge}
Aditya Sanghi, Hang Chu, Joseph~G Lambourne, Ye Wang, Chin-Yi Cheng, Marco
  Fumero, and Kamal~Rahimi Malekshan.
\newblock {CLIP}-{F}orge: Towards zero-shot text-to-shape generation.
\newblock In {\em Proceedings of the IEEE/CVF Conference on Computer Vision and
  Pattern Recognition}, pages 18603--18613, 2022.

\bibitem{schonberger2016structure}
Johannes~L Schonberger and Jan-Michael Frahm.
\newblock Structure-from-motion revisited.
\newblock In {\em Proceedings of the IEEE/CVF conference on computer vision and
  pattern recognition}, pages 4104--4113, 2016.

\bibitem{facenet}
Florian Schroff, Dmitry Kalenichenko, and James Philbin.
\newblock Face{N}et: {A} unified embedding for face recognition and clustering.
\newblock In {\em Proceedings of the IEEE/CVF conference on computer vision and
  pattern recognition}, pages 815--823, 2015.




\bibitem{stap_Conditional}
David Stap, Maurits Bleeker, Sarah Ibrahimi, and Maartje ter Hoeve.
\newblock Conditional image generation and manipulation for user-specified
  content.
\newblock {\em arXiv preprint arXiv:2005.04909}, 2020.

\bibitem{anyface}
Jianxin Sun, Qiyao Deng, Qi Li, Muyi Sun, Min Ren, and Zhenan Sun.
\newblock Any{F}ace: {F}ree-style text-to-face synthesis and manipulation.
\newblock In {\em Proceedings of the IEEE/CVF Conference on Computer Vision and
  Pattern Recognition}, pages 18687--18696, 2022.

\bibitem{sea-t2f}
Jianxin Sun, Qi Li, Weining Wang, Jian Zhao, and Zhenan Sun.
\newblock Multi-caption text-to-face synthesis: {D}ataset and algorithm.
\newblock In {\em Proceedings of the ACM International Conference on
  Multimedia}, pages 2290--2298, 2021.

\bibitem{sun2022fenerf}
Jingxiang Sun, Xuan Wang, Yong Zhang, Xiaoyu Li, Qi Zhang, Yebin Liu, and Jue
  Wang.
\newblock {FE}{N}e{RF}: {F}ace editing in neural radiance fields.
\newblock In {\em Proceedings of the IEEE/CVF Conference on Computer Vision and
  Pattern Recognition}, pages 7672--7682, 2022.

\bibitem{tao2020dfgan}
Ming Tao, Hao Tang, Songsong Wu, Nicu Sebe, Xiao-Yuan Jing, Fei Wu, and Bingkun
  Bao.
\newblock {DF-GAN}: {D}eep fusion generative adversarial networks for
  text-to-image synthesis.
\newblock {\em arXiv preprint arXiv:2008.05865}, 2020.

\bibitem{vaswani2017attention}
Ashish Vaswani, Noam Shazeer, Niki Parmar, Jakob Uszkoreit, Llion Jones,
  Aidan~N Gomez, {\L}ukasz Kaiser, and Illia Polosukhin.
\newblock Attention is all you need.
\newblock In {\em Advances in Neural Information Processing Systems}, pages
  5998--6008, 2017.


\bibitem{faces-a-la-carte}
Tianren Wang, Teng Zhang, and Brian Lovell.
\newblock Faces a la carte: {T}ext-to-face generation via attribute
  disentanglement.
\newblock In {\em Proceedings of the IEEE/CVF winter conference on applications
  of computer vision}, pages 3380--3388, 2021.


\bibitem{xia2021tedigan}
Weihao Xia, Yujiu Yang, Jing-Hao Xue, and Baoyuan Wu.
\newblock Tedi{GAN}: {T}ext-guided diverse face image generation and
  manipulation.
\newblock In {\em Proceedings of the IEEE/CVF Conference on Computer Vision and
  Pattern Recognition}, pages 2256--2265, 2021.

\bibitem{xu2018attngan}
Tao Xu, Pengchuan Zhang, Qiuyuan Huang, Han Zhang, Zhe Gan, Xiaolei Huang, and
  Xiaodong He.
\newblock Attn{GAN}: {F}ine-grained text to image generation with attentional
  generative adversarial networks.
\newblock In {\em Proceedings of the IEEE/CVF conference on computer vision and
  pattern recognition}, pages 1316--1324, 2018.

\bibitem{ye2021xmcgan}
Hui Ye, Xiulong Yang, Martin Takac, Rajshekhar Sunderraman, and Shihao Ji.
\newblock Improving text-to-image synthesis using contrastive learning.
\newblock {\em arXiv preprint arXiv:2107.02423}, 2021.

\bibitem{yu2018bisenet}
Changqian Yu, Jingbo Wang, Chao Peng, Changxin Gao, Gang Yu, and Nong Sang.
\newblock Bi{S}e{N}et: {B}ilateral segmentation network for real-time semantic
  segmentation.
\newblock In {\em Proceedings of the European conference on computer vision},
  pages 325--341, 2018.

\bibitem{parti}
Jiahui Yu, Yuanzhong Xu, Jing~Yu Koh, Thang Luong, Gunjan Baid, Zirui Wang,
  Vijay Vasudevan, Alexander Ku, Yinfei Yang, Burcu~Karagol Ayan, et~al.
\newblock Scaling autoregressive models for content-rich text-to-image
  generation.
\newblock {\em arXiv preprint arXiv:2206.10789}, 2022.

\bibitem{zhang2017stackgan}
Han Zhang, Tao Xu, Hongsheng Li, Shaoting Zhang, Xiaogang Wang, Xiaolei Huang,
  and Dimitris~N Metaxas.
\newblock Stack{GAN}: {T}ext to photo-realistic image synthesis with stacked
  generative adversarial networks.
\newblock In {\em Proceedings of the IEEE international conference on computer
  vision}, pages 5907--5915, 2017.

\bibitem{zhang2018stackgan++}
Han Zhang, Tao Xu, Hongsheng Li, Shaoting Zhang, Xiaogang Wang, Xiaolei Huang,
  and Dimitris~N Metaxas.
\newblock Stack{GAN}++: {R}ealistic image synthesis with stacked generative
  adversarial networks.
\newblock {\em IEEE transactions on pattern analysis and machine intelligence},
  41(8):1947--1962, 2018.


\bibitem{zhang2021m6-ufc}
Zhu Zhang, Jianxin Ma, Chang Zhou, Rui Men, Zhikang Li, Ming Ding, Jie Tang,
  Jingren Zhou, and Hongxia Yang.
\newblock M6-{UFC}: {U}nifying multi-modal controls for conditional image
  synthesis.
\newblock {\em arXiv preprint arXiv:2105.14211}, 2021.

\bibitem{FFHQ_text}
Yutong Zhou.
\newblock Generative adversarial network for text-to-face synthesis and
  manipulation.
\newblock In {\em Proceedings of the ACM International Conference on
  Multimedia}, pages 2940--2944, 2021.


\bibitem{zhu2019dmgan}
Minfeng Zhu, Pingbo Pan, Wei Chen, and Yi Yang.
\newblock {DM-GAN}: {D}ynamic memory generative adversarial networks for
  text-to-image synthesis.
\newblock In {\em Proceedings of the IEEE/CVF Conference on Computer Vision and
  Pattern Recognition}, pages 5802--5810, 2019.

\end{thebibliography}


\end{document}